\documentclass[10pt,a4paper,final,twoside]{article}
\usepackage[utf8]{inputenc}
\usepackage{amsmath}
\usepackage{amsfonts}
\usepackage{amssymb}
\usepackage{graphicx}
\usepackage{placeins}
\usepackage{float}
\usepackage{empheq}
\usepackage{graphicx}
\usepackage{cite}
\usepackage{bbm}
\usepackage{bm}
\usepackage{authblk}
\usepackage{subcaption}
\usepackage{adjustbox}
\usepackage{hyperref}
\usepackage{multirow,color}
\usepackage{array, xspace}
\usepackage{multirow}
\usepackage{multicol}
\usepackage[left=2cm,right=2cm,top=2cm,bottom=2cm]{geometry}

\title{Interactive Storytelling over Document Collections}
\author[1,3]{Dipayan Maiti\thanks{dipayan@vt.edu}}
\author[2,3]{Mohammad Raihanul Islam \thanks{raihan8@cs.vt.edu}} 
\author[1,3]{\\ Scotland Leman\thanks{leman@vt.edu}}
\author[2,3]{Naren Ramakrishnan\thanks{naren@cs.vt.edu}}
\affil[1]{Department of Statistics, Virginia  Tech, VA, USA}
\affil[2]{Department of Computer Science, Virginia  Tech, VA, USA}
\affil[3]{Virginia Tech, Discovery Analytics Center, Arlington, VA 22203}
\date{}

\begin{document}
\maketitle

\begin{abstract}
Storytelling algorithms aim to `connect the dots' between disparate documents by linking
starting and ending documents through a series of intermediate documents. Existing
storytelling algorithms are based on notions of coherence and connectivity,
and thus the primary way by which users can steer the story construction is via design of
suitable similarity functions. We present an alternative approach to storytelling wherein
the user can interactively and iteratively provide `must use' constraints to preferentially 
support the construction of
some stories over others. The three innovations in our approach are distance measures based
on (inferred) topic distributions, the use of constraints to define sets of linear inequalities
over paths, and the introduction of slack and surplus variables to condition the topic
distribution to preferentially emphasize desired terms over others. We describe experimental
results to illustrate the effectiveness of our interactive storytelling approach over multiple
text datasets.
\end{abstract}

\section{Introduction}
Faced with a constant deluge of unstructured (text) data and an
ever increasing sophistication of our information needs, a significant
research front has opened up in the space of
what has been referred to
as {\it information cartography}~\cite{metromaps1}. The basic objective of this space is to pictorially
help users make sense of information through inference of
visual constructs such as stories~\cite{kdd-storytelling2,sto1,dafna1,dafna2}, threads~\cite{threading2,themedelta,threading1}, and maps~\cite{metromaps2,metromaps3}. By supporting interactions
over such constructs, information cartography systems aim to go beyond
traditional information retrieval systems in
supporting users'
information exploration needs.

Arguably the key concept underlying such cartography is the notion of storytelling,
which aims to `connect the dots' between disparate documents by linking
starting and ending documents through a series of intermediate documents. There are
two broad classes of storytelling algorithms, motivated by their different lineages.
Algorithms focused on news articles~\cite{dafna1,dafna2} aim for {\it coherence} of stories
wherein every document in the story shares an underlying common theme.
Algorithms focused in domains such as intelligence analysis~\cite{anWo}
and bioinformatics~\cite{shahriar-plosone}
must often work with sparse information wherein a common theme is typically absent or at best
tenuous. Such algorithms must leverage
weak links to bridge diverse clusters of documents, and thus
emphasize the construction and traversal of similarity networks.
Irrespective of the motivations behind storytelling, all such algorithms provide limited
abilities for the user to steer the story construction process. There is typically no mechanism
to interactively steer the story construction toward desired story lines
and avoid specific aspects that are not of interest.

In this paper, we present an alternative approach to storytelling wherein
the user can interactively provide `must use' constraints to preferentially
support the construction of
some stories over others. At each stage of our approach, the user can inspect the given story and
the overall document collection, and express preferences to adjust the storyline, either in part
or in overall. Such feedback is then incorporated into the story construction iteratively.

Our key contributions are:
\begin{enumerate}
\item Our interactive storytelling approach can be viewed as a form of `visual to parametric interaction'
(V2PI~\cite{v2pi}) wherein users' natural interactions with documents in a workspace is translated into
parameter-level interactions in terms of the underlying machine learning models (here, topic models).
In particular, we demonstrate how high-level user feedback at the level of paths is translated down to
redefine topic distributions.
\item The underlying mathematical framework for interactive storytelling is a novel combination of
hitherto uncombined components:
distance measures based
on (inferred) topic distributions, the use of constraints to define sets of linear inequalities
over paths, and the introduction of slack and surplus variables to condition the topic
distribution to preferentially emphasize desired terms over others. The proposed
framework thus brings together notions from heuristic search, linear systems of inequalities, and topic models.
\item We illustrate how just a modicum of user feedback can be fruitfully employed to redefine
topic distributions and at the same time severely curtail the search process in navigating large
document collections. Through experimental studies, we demonstrate the 
effectiveness of our interactive storytelling approach over multiple text datasets. 
\end{enumerate}

\section{Motivating Example}
We present an illustrative example of how a storytelling algorithm can be steered toward desired lines of
analysis based on user input. For our purposes, assume a vanilla storytelling algorithm
(akin to~\cite{sto1,sto2}) based on heuristic search to prioritize the exploration of
adjacent documents in order to reach a desired destination document. Adjacency here can be
assessed in many ways.
One approach is to use local representations such as a tf-idf representation and define similarity
measures (e.g., Jaccard coefficient) over such local representations.
A second approach is to utilize
the normalized topic distribution generated using, e.g., LDA~\cite{lda}, to induce a distance between 
every pair of documents.

Let us construct a toy corpus of $50$ documents wherein the terms are drawn from $9$ 
predefined 
\textit{themes} and some random \textit{noise} terms. 
Each theme is assumed to be represented by a collection of $4$ terms. An example of a theme is:\\
\textit{Theme 1}: \textbf{nation, terror, avert, orange}\\
%\textit{Theme 3}: \textbf{hazardous, abandoned, sweet, smell}\\
%\textit{Theme 5}: \textbf{ski, tourist, destination, winter}\\
%\textit{Theme 7}: \textbf{bank, red, truck, aspen}\\
%\textit{Theme 8}: \textbf{chemical, factory, recently, hiring}. \\
\noindent
Each document is generated by a single theme or by
mixing two themes. In addition to the terms sampled from the themes, each
document is assumed to also
contain $2$ noise terms. (The noise terms are document-specific meaning two documents do not share the same terms.) 
Thus, we obtain $4$ terms for each of the $9$ themes and $2$ noise terms for each of the $50$ documents, so
that the total number of terms is
$9 \times 4 + 50 \times 2=136$. A pair of documents has an edge between them if they have at 
least one common term. (Since noise terms are not common between the documents, they are not responsible for 
edge formation.) We use the notation $d_n(p\cdots q)$ to denote a document. Here $n$ denotes the 
document index and $p,q$ are the two themes represented by the document. For example $d_1(5\cdots 6)$ is the first document in the corpus and contains terms from themes $5$ and $6$.

%\begin{figure}[!t]
%  \centering
%  \begin{tabular}{c@{}}
%  \includegraphics[width=0.45\textwidth]{fig/story1.pdf} \\
%  (a) \\
%  \includegraphics[width=0.45\textwidth]{fig/story2.pdf} \\
%  (b) \\
%  \includegraphics[width=0.45\textwidth]{fig/story3.pdf} \\
%  (c) \\
%  \end{tabular}
%  \caption{Example}
%  \label{fig:story}
%\end{figure}

\begin{figure}[!ht]
    \centering
    \begin{subfigure}[b]{\textwidth}
            \centering
            \includegraphics[width=0.8\textwidth]{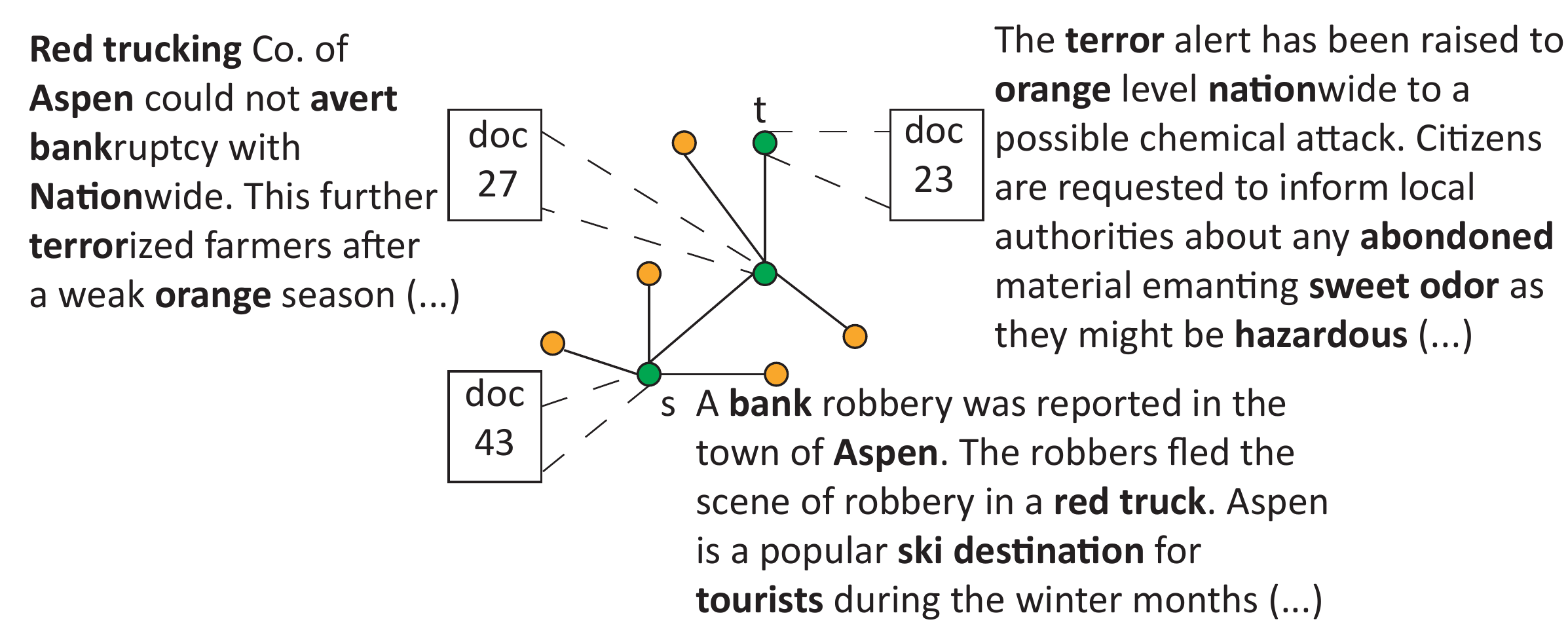}
            \caption{}
            %\caption{User specifies a starting document $s$ reporting a bank robbery and an ending document $t$ alerting about a possible chemical attack. The original \textit{Storytelling} algorithm generates a story which connects the two documents via a document that talks about bankruptcies because of insufficient orange production. The user is not satisfied with this story. }
    \label{fig:sto1}
    \end{subfigure}
	\begin{subfigure}[b]{\textwidth}
            \centering
            \includegraphics[width=0.8\textwidth]{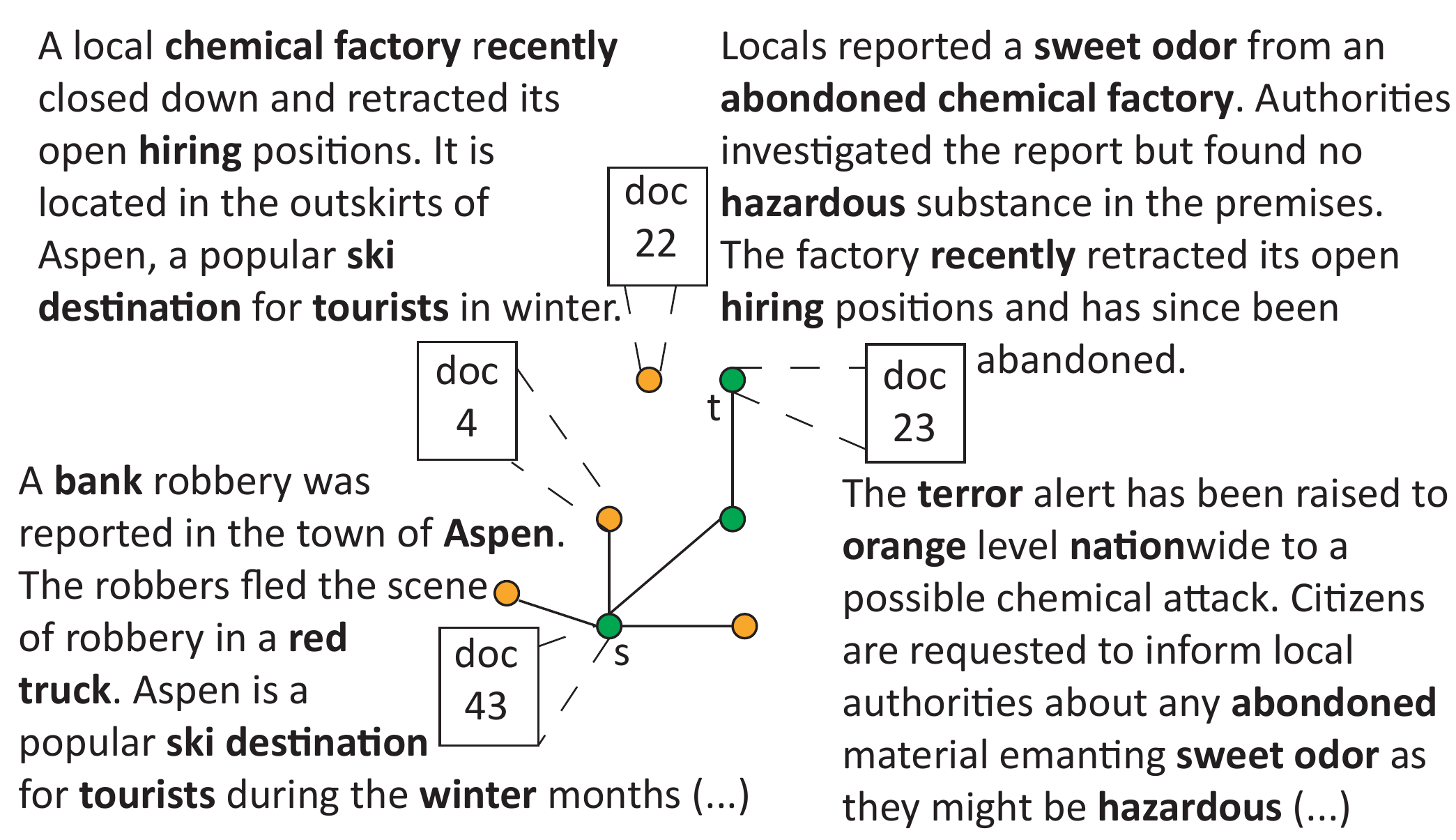}
            \caption{}
            %\caption{User provides feedback by specifying two documents (blue circles) which she expects to be in the story. The first one reports about closure of a chemical factory, and the second one refers to a sweet odor related to chemical weapons emanating from a closed chemical factory.}
    \label{fig:sto2}
    \end{subfigure}
    \begin{subfigure}[b]{\textwidth}
            \centering
            \includegraphics[width=0.8\textwidth]{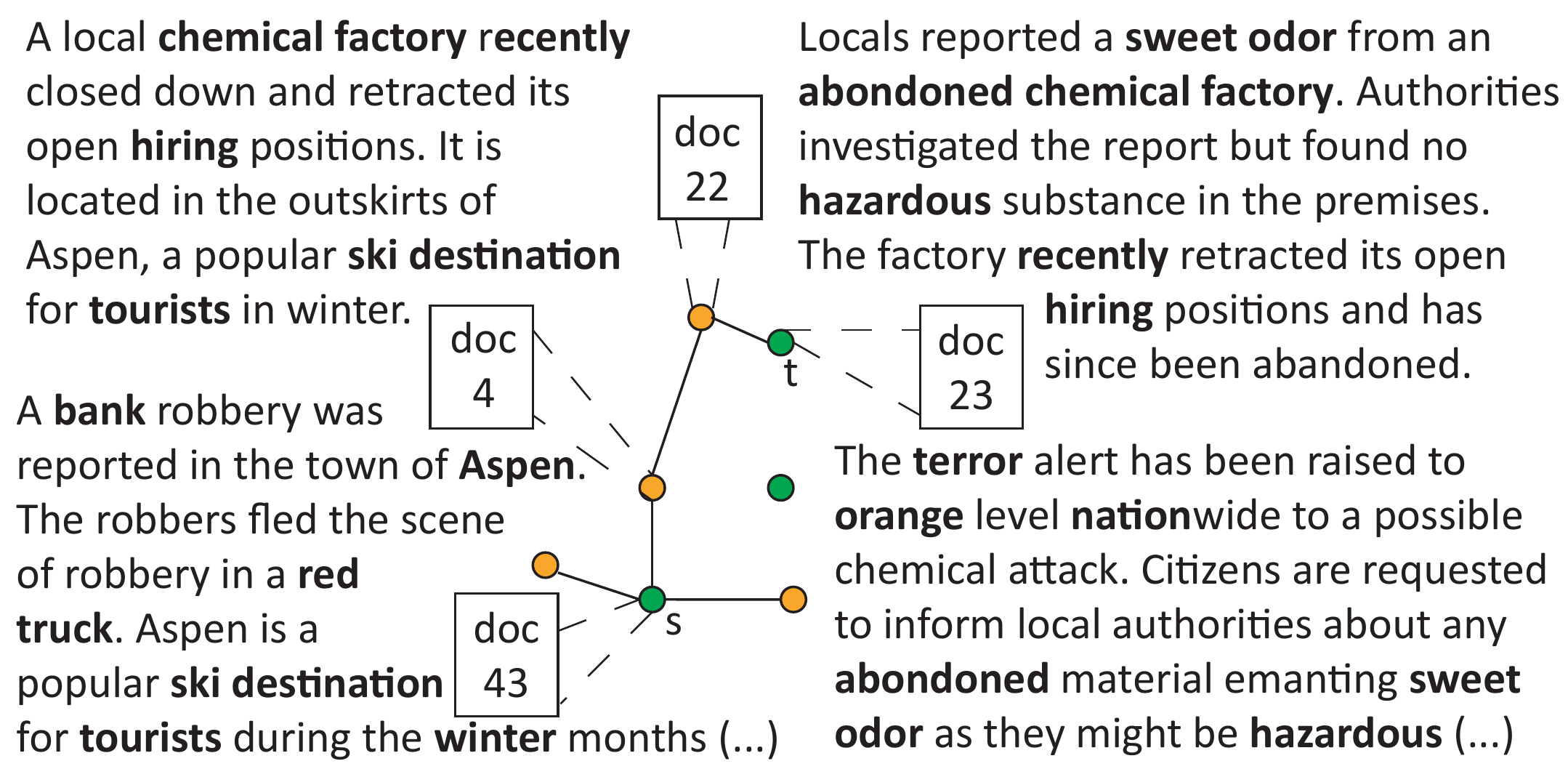}
    \caption{}
    %\caption{Story after incorporating user's feedback based on \textit{Interactive Storytelling}. The first two documents are linked based on mentioning of Aspen, the next two documents are based on abandonment of chemical factory, and the last two are based on a typical odor from chemical weapons}
    \label{fig:sto3}
    \end{subfigure}
    \caption{An illustration of the interactive storytelling algorithm.}
    \label{fig:story}
    %\caption{\protect\subref{fig:fig1} shows figure 1 and \protect\subref{fig:fig2} shows figure 2.}
\end{figure}

Now consider the storytelling scenario from
Fig.~\ref{fig:story}. The user desires to make a story
from document $d_{43}(5\cdots 7)$ to document $d_{23}(1\cdots 3)$. $d_{43}(5\cdots 7)$ describes a bank robbery and 
$d_{23}(1\cdots 3)$ mentions a possible chemical attack. 
The constructed story is as
follows: $d_{43}(5\cdots 7)\rightarrow d_{27}(1\cdots 7)\rightarrow d_{23}(1\cdots 3)$ using
heuristic search (Fig.~\ref{fig:story} (a)).
The first two documents are connected using 
(\textit{Theme 7}), involving the
terms \textbf{bank, red, truck, aspen}.
As can be seen this story is not desirable since the
algorithm has conflated a bank robbery in 
Aspen using a red truck with the bankruptcy of the Red Trucking company 
(due to insufficient orange production in Aspen). Thus although the connection between two documents are 
established by the same set of terms, the contexts are very different. 

In this case the user realizes that the story does not make very good sense, and thus
uses her domain knowledge to steer the story in the right direction. She aims 
to incorporate a story segment $<d_4(5\cdots 8), d_{22}(1\cdots 8)>$ into the
construction. Here, $d_4(5\cdots 8)$ reports the closing of a chemical factory and $d_{22}(1\cdots 8)$ mentions 
about a sweet odor emanating from an abandoned chemical factory (see Fig.~\ref{fig:story} (b)). The user believes that these two documents could
play an important role in the final story. Incorporating this feedback, a story from
$d_{43}$ to $d_{23}$ could potentially
be $d_{43}(5\cdots 7) \rightarrow d_4(5\cdots 8) \rightarrow d_{22}(1\cdots 8) \rightarrow d_{23}(1\cdots 3)$
(i.e., the shortest path from $d_{43}(5\cdots 7)$ to $d_{23}(1\cdots 3)$ via $d_4(5\cdots 8)$ and $d_{22}(1\cdots 8)$). 
Note that there could be other documents necessary to be included in the path 
that are not explicitly provided in the user's feedback.

Incorporating this feedback, the algorithm introduced in this paper
will infer new topic definitions
over the dictionary of terms, and subsequently new
topic distributions for each
document.
In this case, a new story is generated: $d_{43}(5\cdots 7) \rightarrow d_4(5\cdots 8) \rightarrow d_{22}(1\cdots 8) \rightarrow d_{23}(1\cdots 3)$. In this story (see Fig.~\ref{fig:story} (c)), the first two documents are connected by the terms \textbf{ski, tourist, destination, winter} (\textit{Theme 5}); the second and the third are linked via the terms \textbf{chemical, factory, recently, hiring} (\textit{Theme 8}) and the last two documents are connected by \textbf{nation, terror, avert, orange} (\textit{Theme 1}). This story thus suggests an alternative hypothesis for the user's scenario.

\begin{figure}[hbtp]
\centering
\includegraphics[width = 0.8\textwidth]{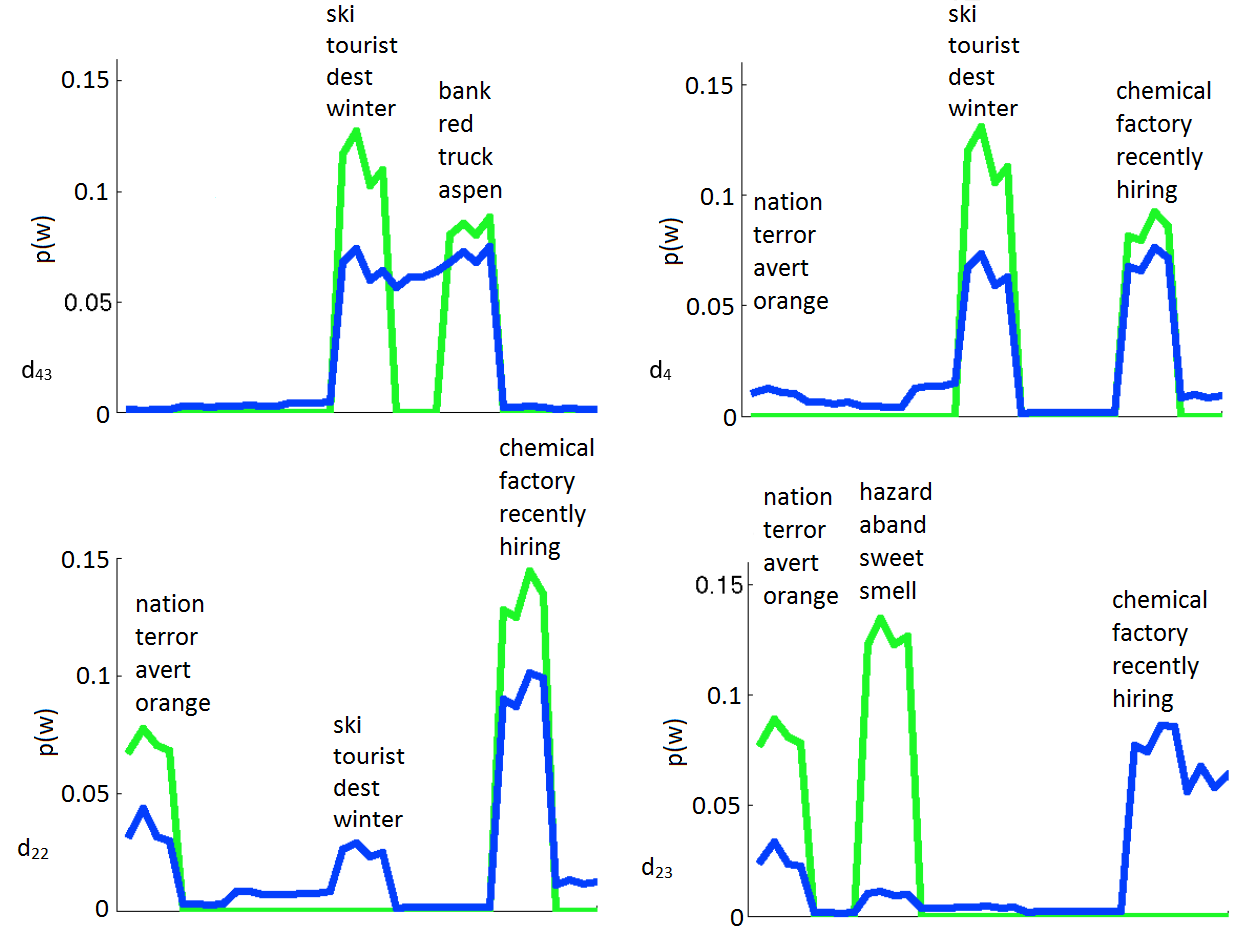}
\caption{Probability of weights of terms before (green) and after (blue) feedback. The inferred topic distributions
are shifted to induce proximity between documents so that the story is consistent with user feedback.} 
\label{fig:mass}
\end{figure}

After incorporating the user's
feedback using our proposed algorithm, we see that \textbf{ski, tourist, destination, winter} has some mass for document $d_{22}(1\cdots 8)$ so that it is inferred
closer to document $d_4(5\cdots 8)$ (see Fig.~\ref{fig:mass}). Similarly, the algorithm estimates positive probabilities for the terms \textbf{chemical, factory, recently, hiring} in document $d_{23}(1\cdots 3)$ which brings it closer to document $d_{22}(1\cdots 8)$. 
%In both cases, the documents may not have the terms but the distribution to the corresponding topic gets mass because of the feedback imposed by the user.

\section{Framework}
A summary of notation as used in this paper is given in Table~\ref{table:sym}.
We utilize
the terms \textit{nodes} and \textit{documents} interchangeably in this paper. 
As described earlier, we impute the notion of distance between
documents based on
vector representations inferred from probabilistic topic models (here, LDA).
Specifically, we use the topic distributions $\theta^{(d_i})$ and $\theta^{(d_j)}$ for documents $d_i$ and $d_j$ (resp.)
to calculate the distance or edge cost between $d_i$ and $d_j$. We posit an edge between two documents if 
they share any terms and the edge cost is lower than a fixed cost $\xi$. 
While a number of probabilistic measure of distance can be utilized,
in this paper we adopt the Manhattan distance metric. The heuristic distance for a node $m$ is 
given by the straight line distance to the ending (target)
document $t$. Since the Manhattan distance obeys the triangle inequality, it is well known that
it is an admissble heuristic for A* search.
As is customary, we define a node
evaluation function $fScore(l)$ as the sum of $gScore(l)$ and $hScore(l)$. 

\begin{table}[!t]
\small
\centering
\caption{Notation overview.}
\label{table:sym}
\begin{tabular}{>{\raggedleft\arraybackslash}p{5cm}|p{11cm}}
\hline \hline
Notation & Explanation \\
\hline
$d_i$ & $i^{th}$ document in the copus \\ 
\hline
$T$ & total number of topics \\
\hline
$s$ & starting document \\
\hline
$t$ & ending/goal document \\
\hline
$\xi$ & distance threshold \\
\hline 
$\theta^{(d_i)}=(\theta_1^{(d_i)},\cdots,\theta_T^{(d_1)})$ & $T$ dimensional vector of normalized topic distribution of document $d_i$ \\ 
\hline
$e_{ij}$ & edge between $d_i$ and $d_j$ if they have any term in common \\
\hline
$c(e_{ij})$ & cost between $d_i$ and $d_j$, $c(e_{ij}) = c_{ij}=\sum_{t=1}^T\Delta_{(ij)t}$, where $\Delta_{(ij)t}=\vert \theta_t^{d_i}-\theta_t^{d_j} \vert$ \\
\hline
$P=<s,d_{P(1)},d_{P(2)},\cdots,d_{L-1},t>$ & path $P$ from $s$ to $t$ with $L$ edges, $d_{P(i)}$ is the $i^{th}$ document after $s$ \\ 
\hline
$c(P)$ & $c(P)=\sum_{e_{ij}\in P} c(e_{ij})$ \\
\hline
$P^{*}$ & shortest path from $s$ to $t$ \\
\hline
$d(i,j)$ & cost of the shortest path from $i$ to $j$ \\
\hline
$gScore(l)$ & cost of the shortest path from $s$ to $l$ using $A^*$ search \\
\hline
$hScore(m)$ & the heuristic distance (Manhattan distance) between the node $m$ and the goal node $t$ \\
\hline
$\alpha_{e^*}$ & minimum cost any $e^*\in E-P^*$ is bounded by such that $P^*$ is the shortest path from $s$ to $t$ \\ 
\hline
$\beta_{e^*}$ & maximum cost any $e^*\in P^*$ is bounded by such that $P^*$ is the shortest path from $s$ to $t$ \\
\hline
$d^{e,k}(s,t)$ & cost of the shortest path from $s$ to $t$ with $c(e)=k$ \\
\hline
$c^{e,k}(P)$ & cost of an arbitrary path $P$ with $c(e)=k$ \\
\hline
$d(s,e,t)$ & cost of the shortest path from $s$ to $t$ including an edge $e\in E$  \\
\hline \hline
\end{tabular} 
\end{table}
%\subsection{Interactive Topic Model}
\subsection{Obtaining User Feedback}
After an initial story generated by heuristic search, the user provides a sequence of documents
that ought to be included in the story (i.e.,
between the documents $s$ and $t$). Suppose this sequence is $\mathcal{C}=<C_1, \cdots, C_K>$. The order of the documents is important, since the sequence is a reflection of desired story
progression. We define
the path $P^*$ as a concatenation of
the shortest path between $s$ and $C_1$, followed by
the nodes in $\mathcal{C}$, and finally the
shortest path between $C_K$ and $t$. This process is done in the original LDA-inferred topic space.
We will now undertake a constrained $A^*$ search incorporating the user feedback.

\iffalse
The user can generate the sequence in a number of ways. She can be motivated by reading the documents in $\mathcal{C}$ or looking for specific term in the documents which may be pertinent to connecting $s$ and $t$. There could be several reason to include a document into the story. However we are ignoring these scenarios. We assume that there is visual analytic platform which can be used by the user to generate the sequence $\mathcal{C}$.

We incorporate the feedback in an effort
to search for parameters that will most likely represent a document as a mixture of topics and 
which will be consistent with the provided feedback
feedback in $\Re$. . %In the next subsection we will describe what an \textit{alternate} stories mean and how this will help to achieve our objective.
\fi
\begin{figure}[!htbp]
\centering
\includegraphics[width=\textwidth]{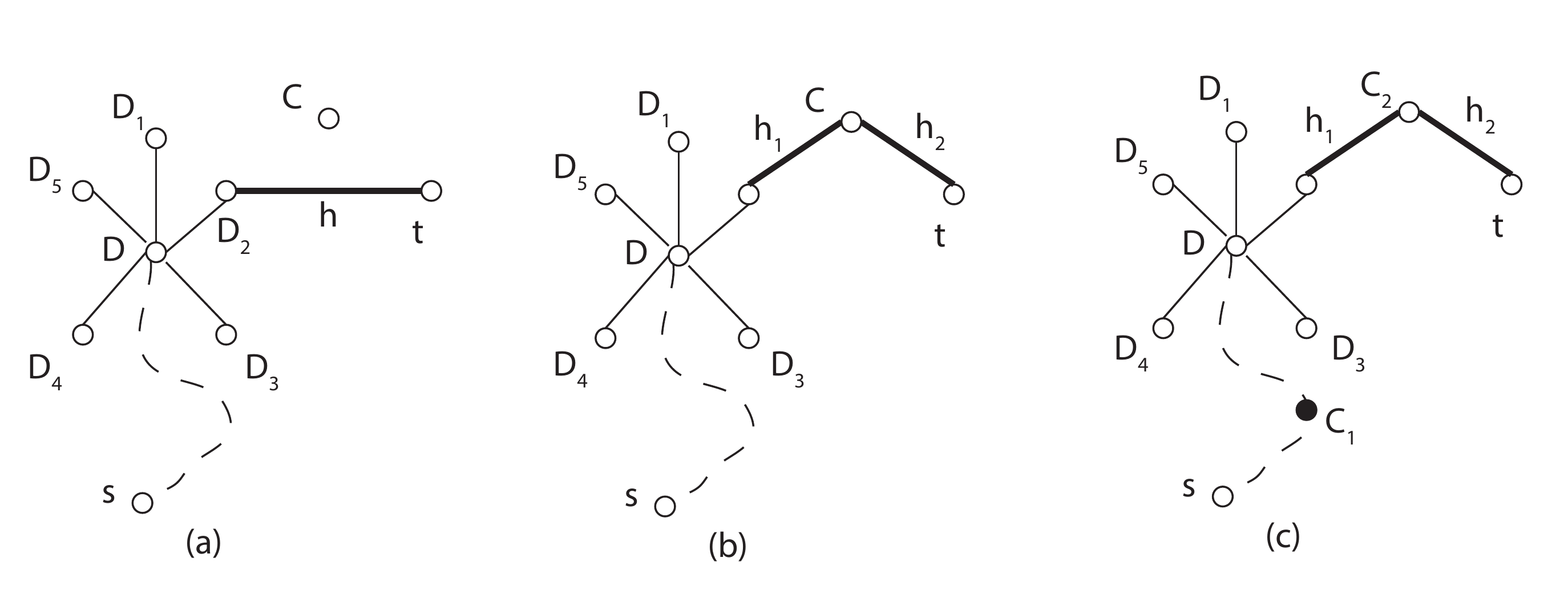}
\caption{(a) shows the heuristic distance $h(D_2,t)$ from original $A^*$ search. (b) depicts $h^*(D_2,t)$ based on constrained $A^*$ search. (c) depicts $h^*(D_2,t)$ when feedback nodes are $\mathcal{C}=<C_1,C_2>$ where ancestry of $D$ is given by $\mathcal{A}(D)=C_1$. Dashed line shows the shortest path from $s$ to $D$.} 
\label{fig:conAStar}
\end{figure}
%\vspace{-0.2cm}
%\vspace{-0.1cm} 
\subsection{Constrained {\large $A^*$} search}\label{sec:cas}
Now we discuss the 
incorporation of the user's feedback
into the story. 
Consider the case where the user insists that a document $C$ (not in the initial story) should be included in 
the story. This case can be easily extended to a sequence of documents $\mathcal{C}=<C_1, \cdots, C_K>$. 
Suppose the adjacent nodes of a document $d$ is denoted by $\mathcal{N}(d)$. There are five adjacent nodes to 
$d$ in Fig.~\ref{fig:conAStar}. The heuristic distance between a neighbor (say, $D_2$) and the ending document $t$ is given by $h(D_2,t)$ in the original $A^*$ search. Our redefined heuristic distance for constrained $A^*$ search is given by $h^*(D_2,B)=h(D_2,C)+h(C,B)$. If the feedback is a sequence of documents $\mathcal{C}=<C_1, \cdots, C_K>$ then $h^*(D_2,t)=h(D_2,C_1)+h(C_1,C_2)+\cdots +h(C_K,t)$. However, while $h^*$ ensures that the $fScore$ of a document depends on the path via the sequence of feedback nodes $\mathcal{C}$, it must also consider the subset of $\mathcal{C}$ that already belong to the shortest path from $s$ to $D$ to estimate the heuristic distance $h^*(D,t)$. We define a property named \textit{Ancestry} that keeps track of the subset of the feedback nodes that already exists in the shortest path from the $s$ to the said node. Ancestry $\mathcal{A}(D_i)$ of an arbitrary neighbor of $D$ is defined as $\mathcal{A}(D_i)=\mathcal{A}(predecessor(D))$ if $D$ is not a feedback node. If $D$ is the feedback node immediately after the subsequence $\mathcal{A}(predecessor(D))$ in $\mathcal{C}$ then $\mathcal{A}(D_i)= \{ (predecessor(D)),D \}$. The starting node $s$ has an empty ancestry. A node having longer subsequence of $\mathcal{C}$ in its ancestry compared to another is said to have a \textit{richer} ancestry. A node with richer ancestry is always preferred. If ancestries are comparable, for an open node the predecessor with smaller $gScore$ is chosen while for a closed node the predecessor with smaller $fScore$ is chosen.

\begin{figure}[!t]
\centering
\includegraphics[width=\textwidth]{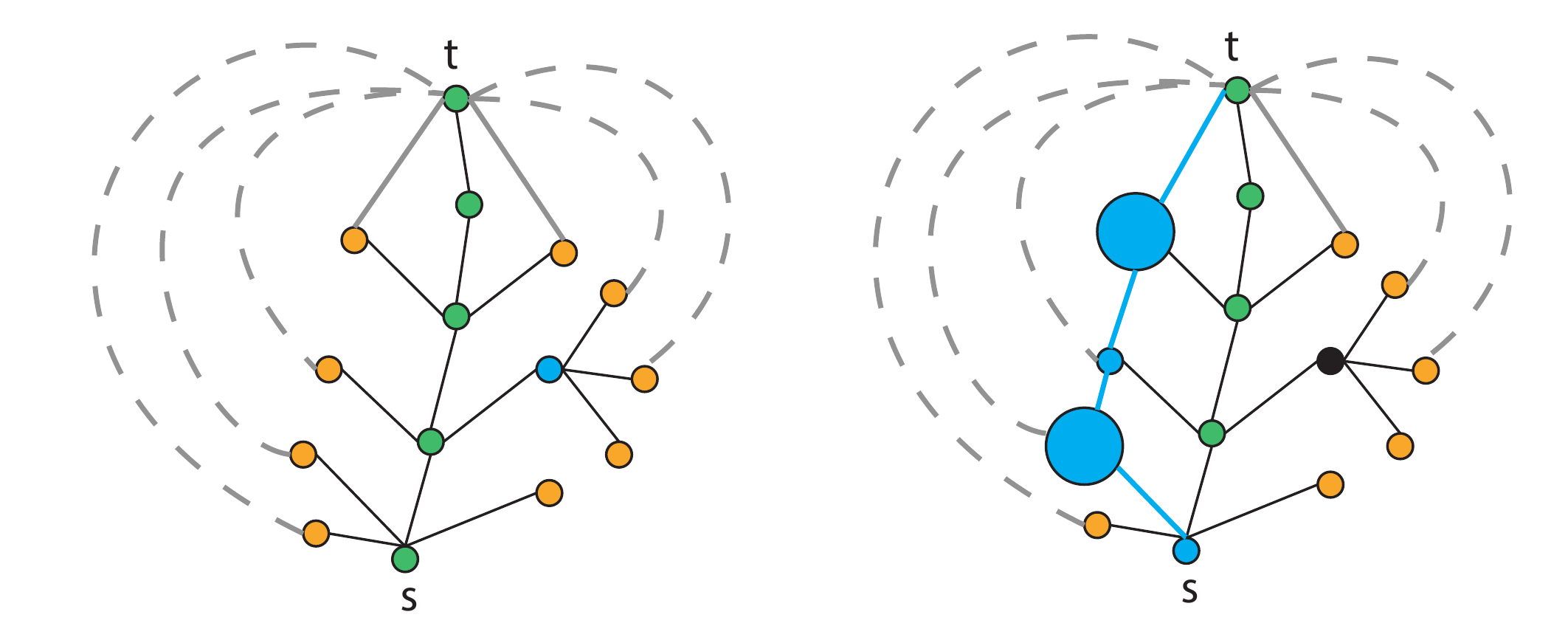}
\caption{(left)
The path with green nodes is the initial story generated by the storytelling
algrithm
and hence the shortest path from $s$ to $t$ before incorporating feedback. The gray paths (dashed and solid) are alternate stories abandoned by the $A^*$ search. (right)
Story after user feedback where
the user-preferred story $P^*$ is shown
in blue. This is not the shortest path in the current topic space. The documents that the user desires to 
be in the story are shown in large circles. We intend to estimate the topic space where the 
blue path ($P^*$) is shorter than all the other alternate paths from $s$ to $t$.}
\label{fig:feedback}
\end{figure}
%\vspace{-0.2cm} 
\subsection{Alternate/Candidate Stories}
The nodes explored by $A^*$ search in the initial topic space (the set of open and closed nodes) induce an acyclic graph $G(V, E)$. The orange nodes in Fig.~\ref{fig:feedback} are open nodes in such a graph. Denote the set of open nodes by $\mathcal{O}$. Any path from $s$ to $t$ via $o \in \mathcal{O}$ is a candidate story generated by $A^*$ search. Let us denote the path via $o$ by $P^{(o)}$.

Now assume $\mathcal{O}$ has $O$ open nodes. To enforce the user feedback that $P^*$ be the shortest path over all paths from $s$ to $t$ we define the following system of inequalities:
%\vspace{-0.1cm} 
\small
\begin{align}\label{eqn:inq}
c(P^*) &\leq c(P^{(o_1)}) \nonumber \\
&\vdots \nonumber \\
c(P^*) &\leq c(P^{(o_O)}) 
\end{align}
\normalsize
If we break each inequality in terms of topics then we obtain:
%\vspace{-0.1cm} 
%\small
\begin{align}
\sum_{t=1}^T(\Delta_t^*-\Delta^{(o_1)}) &\leq 0 \nonumber \\
&\vdots \nonumber \\
\sum_{t=1}^T(\Delta_t^*-\Delta^{(o_O)}) &\leq 0 \label{eqn:inqT}
\end{align}
%\normalsize
In addition to this set of inequalities, we also add another set of inequalities imposing that the cost of an edge in the new topic space, $c(e)$ is at least as much as the cost of the edge in the initial topic space $c_0(e)$. 
%\vspace{-0.1cm} 
\small
\begin{equation}\label{eqn:e}
c(e) \geq c_0(e), e \in E
\end{equation}
\normalsize
This constraint is imposed so that the proximity of the document does not change drastically, as otherwise this might
disorient users.
%\vspace{-0.1cm} 
\subsection{Deriving Systems of Inequalities}
$A^*$ is a heuristic algorithm to find the shortest path between two nodes. Given the shortest path, finding the edge costs or upper and lower limits thereof is thus as \textit{inverse shortest path problem}. Our goal is to find a normalized topic distribution $\theta^{(d_i)}$ so that $P^*$ is actually the shortest path in the new topic space.

In our approach, we obtain the inequalities in Eqn \ref{eqn:inqT} by using the following observation: if the cost of an edge $e^*\in P^*$ crosses the upper threshold $\beta_e^*$ or an edge $e \not\in P^*$ falls below the lower threshold $\alpha_e$, all the other edge cost being fixed $P^*$ is no longer the shortest path from $s$ to $t$. Therefore the condition for $P^*$ being the shortest path is 
\small
\begin{align}
c(e^*)\leq \beta_e^*, \forall e^*\in P^* \nonumber\\
c(e)\geq \alpha_e, \forall e \in E-P^*
\end{align}
\normalsize
Upper and lower shortest path tolerances are presented in~\cite{tolSP} as:
\small
\begin{align}\label{eqn:ab}
\beta_{e^*}&=d^{e^*,\infty}(s,t)-c(P^*)+c(e^*) \nonumber\\
\alpha_e&=c(P^*)-d^{e,0}(s,t)
\end{align}
\normalsize
Therefore the inequities for the edges becomes:
\small
\begin{align}\label{eqn:pstar}
c(P^*)&\leq d^{e^*,\infty}(s,t), \forall e^* \in P^* \\ 
c(e)&\geq c(P^*)-d^{e,0}(s,t), \forall e \in E-P^* \label{eqn:eP}
\end{align}
\normalsize
%Now $c(e^*)>\beta$ means $d^{e^*,\infty}(s,t)<C(P^*)$ which is exactly we want to avoid the first in
Note that for the first equation in Eqn.~\ref{eqn:ab}, $\beta_{e^*}$ is the difference of two path costs: the cost of the shortest path from $s$ to $t$ that avoid $e^*$ (imposing an infinite cost for $e^*$) $d^{e^*,\infty}(s,t)$ and the minimum cost of $P^*$ with $e^*$ in the path (imposing a zero cost for $e^*$), so that
$c(P^*)-c(e^*)=c^{e^*,0}(P^*)$. Notice also that
if $e=(l,m)$, then $d^{e,o}(s,t)=\min(c(P^*),d(s,l)+d(m,t))$. For the second equation if the shortest path from $s$ to $t$ does not change even with $c(e)=0$, i.e. $d^{e,0}(s,t)=c(P^*)$, then the lower tolerance for $c(e)$ is zero. However, if the constraint $c(e)=0$ favors a different path through $e$ (meaning not $P^*$) the lower tolerance for $e$ is given by the drop in the path cost which this alternate path allows over $P^*$.

We use the fact that our choice of $hScore$ is an admissible heuristic in $A^*$ search to simplify our formulation of inequalities. Due to admissibility, $hScore(m)\leq d(m,t)$, and
consequently $gScore(l)+hScore(m)\leq d(s,l)+d(m,l)$. Replacing $d^{e,0}(s,t)$ with lower heuristic estimate of $gScore(l)+hScore(m)$ in Eqn. \ref{eqn:eP} we achieve a stricter inequality:
\small
\begin{equation}
\left.\begin{matrix}
c(e)\geq c(P^*)-gScore(l)-hScore(m)\\
c(e)\geq 0
\end{matrix}\right\} \forall e \in E-P^* 
\end{equation}
\normalsize
\begin{figure}[!t]
\centering
\includegraphics[width=0.5\textwidth]{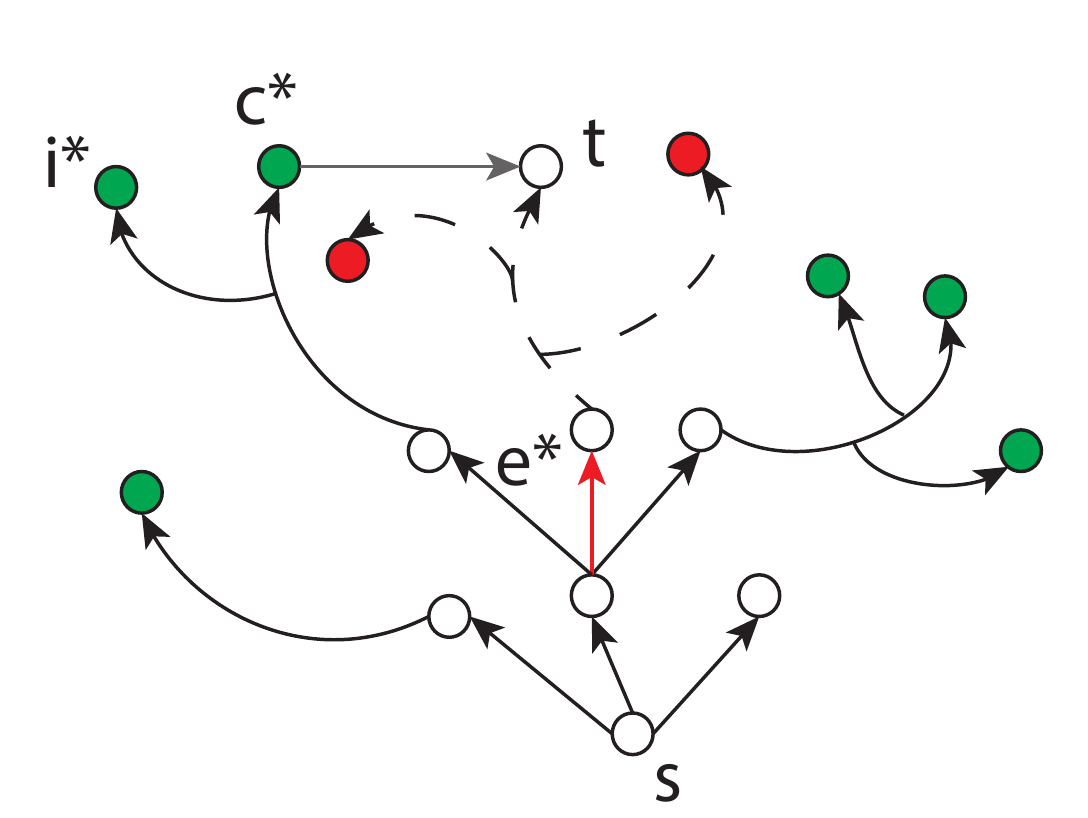}
\caption{Dashed line shows the subtree $\tau(e^*)$ and the solid line shows the subtree $\tau^C(e^*)$. The candidate open nodes in $\tau^C(e^*)$ for Eqn. \ref{eqn:pstar} are shown in green. Red nodes are open nodes in $\tau(e^*)$ and do not contribute in Eqn. \ref{eqn:pstar}. The shortest path from $s$ to $t$ avoiding $e^*$ is the shortest path from $s$ to $t$ via any of the green nodes.}
\label{fig:open}
\end{figure}
The cost of shortest path avoiding $e^*\in P^*$ is given by $d^{e^*,\infty}(s,t)=\min_{e\in E-P^*}(d(s,e,t)|e^*\not\in d(s,e,t))$. In Fig. \ref{fig:open} suppose the red edge is one such $e^*\in P^*$. Let the subtree induced by $A^*$ search following $e^*$ is $\tau(e^*)$ (shown in dashed line) and the remainder of the tree is $\tau^C(e^*)$ (shown in solid line). Based on the search process, we would expect the shortest path from $s$ to $t$ via any edge in $\tau(e^*)$ to have $e^*$ in it. Therefore $d^{e^*,\infty}(s,t)$ should be based on paths via edges in $\tau^C(e^*)$. Since we have path costs that are estimated by the heuristically $A^*$ search ($fScores$) we can use these for the open nodes in $\tau^C(e^*)$. These open nodes are shown in green in Fig.~\ref{fig:open}. Hence in this setting, the inequality $c(P^*)\leq \min_{e\in E-P^*}(d(s,e,t)|e^*\not\in d(s,e,t))$ is replaced by the following set of inequalities:
%\vspace{-0.2cm}
\small
\begin{equation}\label{eqn:o}
c(P^*) \leq fScore(o), \forall o\text{ in the set of open nodes in }\tau^C(e^*)
\end{equation}
\normalsize
%\vspace{-0.05cm}
Due to the admissibility of $hScore$, $fScore$ also underestimates the true distance, so we are using a stricter inequality in Eqn. \ref{eqn:o}. If this process is repeated for all $e^*\in P^*$ our set of inequalities consist of the user defined path $P^*$ being compared against all the set of paths defined by the open nodes in the original $A^*$ search given in Eqn \ref{eqn:inqT}. 
%\vspace{-0.2cm} 
\subsection{Modeling Relationships by Auxiliary Variables}
In the previous section we formulated the user feedback as a set of relationships, where each relationship is an inequality in terms of path lengths. Since the distance metric is based on normalized topic distribution we explicitly show the dependence of an individual relationship on $\pmb{\theta}$. For an inequality $r_o\equiv c(P^*)\leq c(P^{(o)})$  in Eqn. \ref{eqn:inqT}, we introduce a slack random variable $\lambda_o$ (i.e. $\lambda_o\leq\epsilon$ for some $\epsilon\leq 0$) as an auxiliary variable with expectation $\pmb{E}(\lambda_o)=\mu_o(\pmb{\theta})=c(P^*)-c(P^{(o)})$. Similarly for a relationship $r_e\equiv c(e)\geq c_0(e)$ in Eqn \ref{eqn:e} we define a surplus random variable $\lambda_e$ where $\lambda_e$ is positive with expectation given by $\pmb{E}(\lambda_e)=\mu_o(\pmb{\theta})=c(e)-c_0(e)$. Therefore $\mu_o(\pmb{\theta})=\sum_{t=1}^T(\Delta_t^*(\pmb{\theta})-\Delta_t^{(o)}(\pmb{\theta}))$. Suppose the distribution of the auxiliary variable is given by $\lambda_o \sim f(\cdot\vert\pmb{\theta})$. The random variable $\lambda_o$ measures the difference in path lengths between the user defined path $P^*$ and an alternate $P^{(o)}$. If $\mu_o(\pmb{\theta})$ is zero, it means enforcing the relationship that $P^*$ is as costly as the alternate path $P^{(o)}$. The more negative the value of its mean $\mu_o(\pmb{\theta})$, the larger we expect $P^{(o)}$ to be compared to $P^*$. This ensures that the topic space $\pmb{\theta}$ satisfies the relationship $c(P^*) \leq c(P^{(o)})$. Now conditional on a known $\pmb{\theta}$, the joint distribution of the auxiliary variables (both slack and surplus) and the observed feedback $\Re$ is given below:
%\vspace{-0.1cm}
\begin{equation}\label{eqn:jd}
\begin{split}
f(\Re,\pmb{\lambda}|\pmb{\theta}) &\propto \prod_{o\in \mathcal{O}}\{ \mathbbm{1}_{c(P^*) \leq c(P^{(o)}}\mathbbm{1}_{\lambda_o\leq\epsilon} +\mathbbm{1}_{c(e)\geq c_o(e)}\mathbbm{1}_{\lambda_0\geq 0} \}f(\lambda_o|\pmb{\theta})
\end{split}
\end{equation}
%\vspace{-0.1cm}
Here, $\mathbbm{1}_x$ is an indicator variable which is one if condition $x$ holds and zero
otherwise. Our goal is to find a set of surplus and slack variables $\pmb{\lambda}$ that maximizes the probability in Eqn \ref{eqn:jd}. Now let $f(\lambda_o|\pmb{\theta})$ be normally distributed with mean $\mu_0(\pmb{\theta})$ and variance 1. By marginalizing over the auxiliary variables $\lambda_o$, our formulation is same as the modeling the probability of satisfying a relationship using the cumulative normal distribution. 
%\vspace{-0.2cm}
\small
\begin{align}
P(c(P^*) \leq c(P^{(o)})|\pmb{\theta}) &= 1 - \Phi(\mu_o(\pmb{\theta})-\epsilon), \text{for Eqn \ref{eqn:inqT}} \nonumber \\
P(c(e)\geq c_0(e)|\pmb{\theta}) &= \Phi(\mu_o(\pmb{\theta})), \text{for Eqn \ref{eqn:e}}
\end{align}
\normalsize
%\vspace{-0.1cm}
Here for a standard normal variable $Z$, $\Phi(z)=P(Z\leq z)$. This approach is very similar to the usage of auxiliary variables in probit regression \cite{probit}. In probit regression the mean of the auxiliary variable is modeled by a linear predictor to maximize the discrimination between the successes and failures in the data. In our case satisfiability of a user defined relationship is a success and the probability of satisfying the relationship is modeled by the mean of auxiliary variable. The mean of the auxiliary variable is a function of the topic space $\pmb{\theta}$ on which the distances are defined. Our goal is to search for a topic space $\pmb{\theta}$ which explains the term distribution of the documents and satisfies as many of the relationships in $\Re$ as possible. Truncating a slack variable $\lambda_o$ to a negative region specified by $\epsilon$ allows to search for $\pmb{\theta}$ that shrinks the mean $\mu_o(\pmb{\theta})$ to a negative value. The complete hierarchical model using the term document data $\pmb{\eta}$ and the relationship data $\Re$ is presented below:
%\vspace{-0.3cm}
%\small
\begin{align}
\begin{split}
f(\Re,\pmb{\lambda}|\pmb{\theta}) &\propto \prod_{o\in \mathcal{O}}\{\mathbbm{1}_{c(P^*) \leq c(P^{(o)}}\mathbbm{1}_{\lambda_o\leq\epsilon} +\mathbbm{1}_{c(e)\geq c_o(e)}\mathbbm{1}_{\lambda_o\geq 0} \}N(\lambda_o|\mu_o(\pmb{\theta}),1)
\end{split} \nonumber \\
\eta_i|z_i,\phi^{(z_i)} &\sim Discrete(\phi^{(z_i)}) \nonumber \\
\phi &\sim Dirichlet(\beta) \nonumber \\
z_i|\theta^{(d_i)} &\sim Discrete(\theta^{(d_i)}) \nonumber \\
\theta &\sim Dirichlet(\alpha)
\end{align}
%\normalsize
%\vspace{-0.5cm} 
\subsection{Inference}
We use Gibbs sampling to compute the posterior distributions for $\mathbf{z}, \pmb{\lambda}$ and $\pmb{\theta}$. The conditional posterior distributions for $z_i$ is given below:
\small
\begin{equation}\label{eqn:sampT}
p(z_i=j|\mathbf{z}_{(-i)},\pmb{\eta})\propto p(\eta_i|z_i=j,\mathbf{z}_{(-i)},\pmb{\eta}_{(-i)})p(z_i = j|\mathbf{z}_{(-i)},\pmb{\eta}_{(-i)})
\end{equation}
\normalsize
%The first term in Eqn \ref{eqn:sampT} can be calculated by marginalizing over 
The sampling of topic for terms $\pmb{\eta}$ is same as used in vanilla LDA~\cite{pnastm}.
%\vspace{-0.1cm} 
\small
\begin{equation}
p(z_i = j | \mathbf{z}_{(-i)},\pmb{\eta}) \propto \dfrac{\beta+n^{(\eta_i)}_{(-i,j)}}{M\beta+n^{(\cdot)}_{(-i,j)}} \times \dfrac{\alpha+n_{(-i,j)}^{(d_i)}}{T\alpha + n_{(-i,\cdot)}^{(d_i)}}
\end{equation}
\normalsize
The full conditional distribution for $\lambda_o$ is given below:
\small
\begin{equation}
p(\lambda_o | \pmb{\theta},\Re) = \begin{cases}
N(\cdot|\mu_o(\pmb{\theta}),1),\lambda_o\leq \epsilon, \text{if } r_o \text{ is } \leq \text{ type}\\
N(\cdot|\mu_o(\pmb{\theta}),1),\lambda_o>0, \text{if } r_o \text{ is } > \text{ type}
\end{cases}
\end{equation}
\normalsize
The full conditional distribution for the topic distribution of document $d_j$ is given below:
\small
\begin{align}
\begin{split}
p(\theta^{(d_j)}|\pmb{\theta}^{(-d_j)},\pmb{\lambda},\pmb{z}) &\propto \prod_{z_i\in d_j}p(z_i|\theta^{(d_j)})p(\theta^{(d_j)}|\alpha)\times \prod_{o\in \mathcal{O}}N(\lambda_o|\mu_o(\pmb{\theta}),1) 
\end{split}\nonumber \\
&\propto p(\theta^{(d_j)}|\pmb{z},\alpha)\prod_{o\in \mathcal{O}}N(\lambda_o|\mu_o(\pmb{\theta}),1) \nonumber \\
&\propto \prod_{t=1}^T \left( \theta_t^{(d_j)} \right)^{(n_t^{(d_j)}+\alpha)-1}\prod_{o\in\mathcal{O}}N(\lambda_o|\mu_o(\pmb{\theta}),1)
\end{align}
\normalsize
since $p(\theta^{(d_j)}=Dirichlet(n_t^{(d_j)}+\alpha)$. $n_t^{(d_j)}$ denotes the number of terms in document $d_j$ assigned to topic $t$ based on $\pmb{z}$. If $d_j$ does not belong to $\Re$, then $\theta^{(d_j)}$ is sampled from $Dirichlet(n_t^{(d_j)}+\alpha)$. We sample from $p(\theta^{(d_j)}|\pmb{\theta}^{(-d_j)},\pmb{\lambda},\pmb{z})$ by a Metropolis-Hastings step otherwise. We use a proposal strategy based on stick-breaking process to allow better mixing. The stick-breaking process bounds the topic distribution of a document $d_j$ between zero and one and their sum to one. We first sample random variables $u_1,\cdots,u_{T-1}$ truncated between zeros and one and centered by $\pmb{\theta}^{(d_j)}$ using a proposal distribution $q(\cdot)$:
\small
\begin{align}
u_1&\sim q\left(\cdot|\theta_1^{(d_j)}\right), 0<u_1<1 \nonumber \\
u_2&\sim q\left(\cdot|\dfrac{\theta_2^{(d_j)}}{1-u_1}\right), 0<u_2<1 \nonumber \\
u_3&\sim q\left(\cdot|\dfrac{\theta_3^{(d_j)}}{(1-u_1)(1-u_2)}\right), 0<u_3<1 \nonumber \\
\vdots \nonumber \\
u_{T-1}&\sim q\left(\cdot|\dfrac{\theta_{T-1}^{(d_j)}}{(1-u_1)(1-u_2)\cdots (1-u_{T-2})}\right), 0<u_{T-1}<1
\end{align}
\normalsize
%\vspace{-0.1cm}
% u_3&\sim q\left(\cdot|\dfrac{\theta_3^{(d_j)}}{(1-u_1)(1-u_2)}\right), 0<u_3<1 \nonumber \\
This is followed by the mappings, $S:\mathbf{u}\rightarrow \pmb{\theta}_{1:T-1}^{*(d_j)}$,
%\vspace{-0.2cm}
\small
\begin{align}
\theta_1^{*(d_j)} &= u_1 \nonumber \\
\theta_2^{*(d_j)} &= u_2(1-u_1) \nonumber \\
\theta_3^{*(d_j)} &= u_3(1-u_2)(1-u_1) \nonumber \\
\vdots \nonumber \\
\theta_{T-1}^{*(d_j)} &= (1-u_{T-1})(1-u_{T-2})\cdots (1-u_2)(1-u_1)
\end{align}
\normalsize
The inverse mappings $S^{-1}:\pmb{\theta}_{1:T-1}^{*(d_j)}\rightarrow \mathbf{u}$ are given by:
\small
\begin{align}
u_1&=\theta_1^{*(d_j)} \nonumber \\
u_t &= \dfrac{\theta_1^{*(d_j)}}{1-\sum_{i<t}\theta_i^{*(d_j)}}, t=2,\cdots T-1
\end{align} 
\normalsize
The Metropolis-Hastings acceptance probability for such a proposed move is given by 
\small
\begin{equation}
\begin{split}
p_{MH}&=\min \left(1, \dfrac{(p(\theta^{*(d_j)})|\mathbf{z})\prod_{o\in \mathcal{O}}N(\lambda_o|\mu_o(\pmb{\theta}^*),1))}{(p(\theta^{(d_j)})|\mathbf{z})\prod_{o\in \mathcal{O}}N(\lambda_o|\mu_o(\pmb{\theta}),1))} \times \right. \left. \dfrac{q(\pmb{\theta}^{*(d_j)}_{1:T-1})}{q(\mathbf{u})} \left| \dfrac{\delta(\pmb{\theta}^{*(d_j)}_{1:T-1})}{\delta(\mathbf{u})} \right| \right)
\end{split}
\end{equation}
\normalsize
where 
\small
$\left| \dfrac{\delta(\pmb{\theta}^{*(d_j)}_{1:T-1})}{\delta(\mathbf{u})} \right|=\left| \dfrac{\delta(\mathbf{u})}{\delta(\pmb{\theta}^{*(d_j)}_{1:T-1})} \right|^{-1}$=$\left( \dfrac{1}{\prod_{t=2}^{T-1}\left( 1-\sum_{i<t}\theta_i^{*(d_j)} \right)} \right)$
\normalsize
The samples from $\mathbf{z},\pmb{\lambda}$ and $\pmb{\theta}$ are iteratively sampled to generate the joint posterior distribution of all the unknown parameters using Gibbs Sampling. 

This procedure completes the interactivity loop in the storytelling algorithm.
The newly inferred topic distributions will induce a new similarity network over which
we can again conduct a search, followed by (potentially) additional user feedback.

%\vspace{-0.3cm}
\section{Experimental Results}
We evaluate our interactive storytelling approach over a range of text datasets from intelligence analysis,
such as {\it Atlantic Storm}, {\it Crescent}, {\it Manpad}, and the {\it VAST11} dataset from the IEEE Visual
Analytics Science \& Technology Conference. Pl see~\cite{hao-paper} for details of these datasets.
The questions we seek to answer are:
%\vspace{-0.2cm}
\begin{enumerate}
\item Can we effectively visualize the operations of the interactive storytelling as user feedback
is incorporated? (Section \ref{sec:visIn})
%\vspace{-0.2cm}
\item Does the interactive storytelling framework provide better alternatives for stories than
a vanilla topic model? (Section \ref{sec:ks})
%\vspace{-0.2cm}
\item Are topic reoorganizations obtained from interactive storytelling
significantly different from a vanilla topic model? (Section \ref{sec:proT})
%\vspace{-0.2cm}
\item Does our method scale to large datasets? (Section \ref{sec:clusS})
%\vspace{-0.2cm}
\item How effectively does the interactive storytelling approach 
improve over uninformed search (e.g., uniform cost search or breadth-first search)? (Section \ref{sec:com})
\end{enumerate}
%\noindent
In the below, unless otherwise stated, we fix the number of topics to be
$T=20$ and set $\alpha=0.05/T$ and $\beta=0.01$. We also use the 
Gini index to remove top 10\% of of the terms as a pre-processing step for our text
collections.

%\begin{table}
%\centering
%\begin{tabular}{c|c}
%\hline \hline
%Name & $\#$ of documents \\ 
%\hline 
%Atlantic Storm & 111 \\ 
%Crescent & 41 \\ 
%Manpad & 47 \\ 
%Vast & 4474 \\
%\hline \hline
%\end{tabular} 
%\caption{Dataset Statistics}
%\label{table:ds}
%\end{table}
%\subsection{Dataset Description}
%\vspace{-0.2cm}
\subsection{Visualizing interactive storytelling} 
\label{sec:visIn}
\begin{figure*}[!t]
\centering
\includegraphics[width=0.95\textwidth]{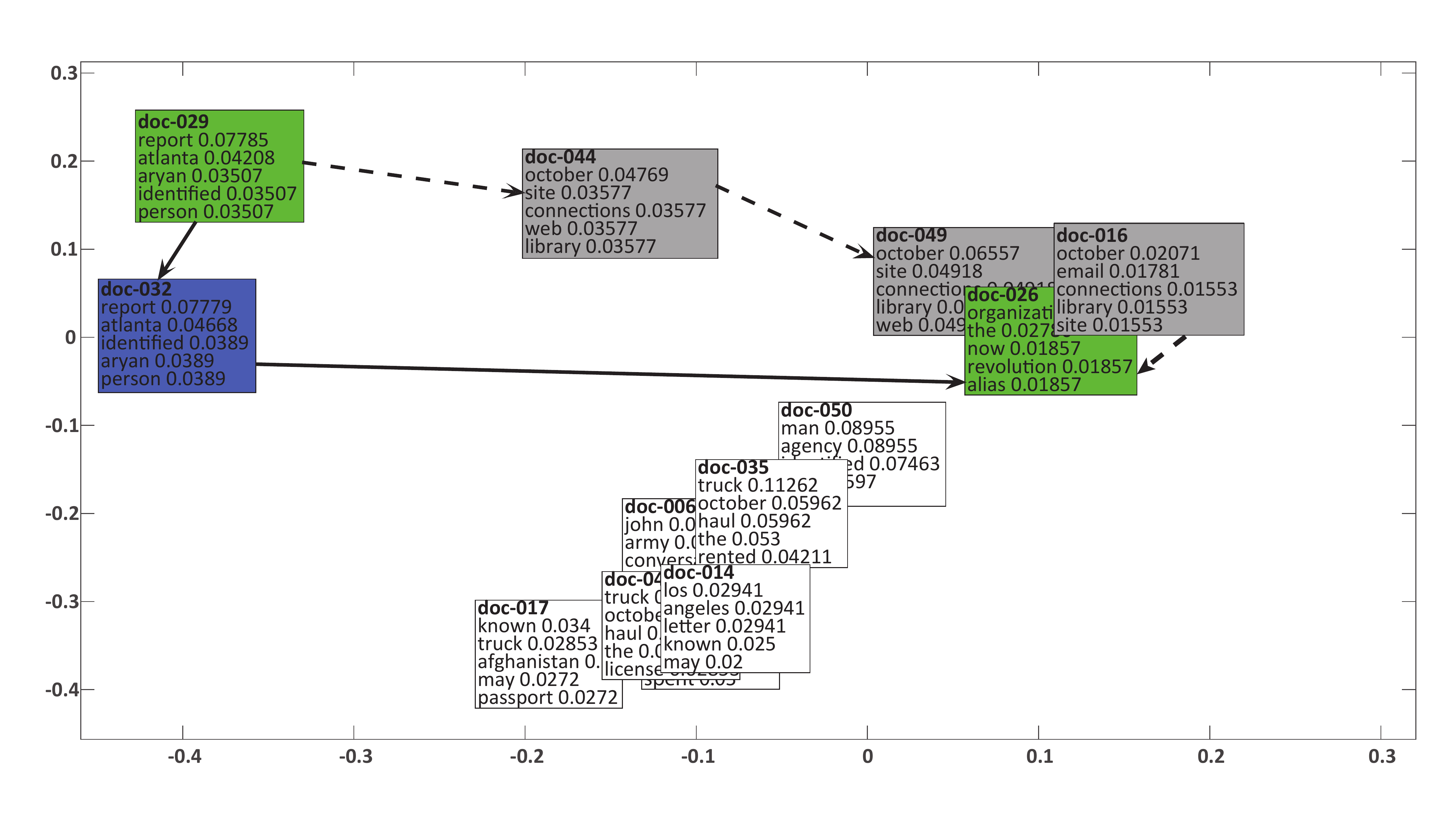}
\caption{Visualizing documents using multidimensional scaling (Manpad dataset) before and after
user feedback. Many documents are omitted for better visualization. The starting and the ending documents are 
shown in green. The documents in the initial story are shown in blue (and the story by solid
lines).  The story generated by the interactive storytelling algorithm is shown in the dotted line through
the grey documents. Each document is represented by its top five terms having the highest posterior probability.}
\label{fig:mds_mand}
\end{figure*}

We apply multidimensional scaling (MDS) over the normalized topic 
space as an aid to visualize the operations of the storytelling
algorithm. For instance, the Manpad dataset is visualized as shown in
Fig.~\ref{fig:mds_mand}. Consider a story from document $Doc$-$29$ to document $Doc$-$26$.
Here $Doc$-$29$ reports that a member of an infamous terrorist organization has a meeting with 
a notorious arms dealer. $Doc$-$26$ reports that a team of suicide bombers plans to set off 
bombs in trains carrying tens of thousands of commuters under the Hudson River. The storytelling 
algorithms generates a story as: $Doc$-$29\rightarrow Doc$-$32\rightarrow Doc$-$26$.  Here,
$Doc$-$32$ identifies a person belong to a terrorist organization. The user is not satisfied with this story 
and provides a constraint that the story should involve
documents $Doc$-$44$ and $Doc$-$49$. Here, $Doc$-$44$ describes that libraries in Georgia and Colorado have
some connections to a web site. $Doc$-$49$ reports that a code number is found in the website linked to a charitable organization. 
Using this feedback a new story is generated:  $Doc$-$29\rightarrow Doc$-$44\rightarrow Doc$-$49\rightarrow Doc$-$16\rightarrow Doc$-$26$.  In addition to being consistent with the user's feedback, note that the algorithm has introduced
a new document ($Doc$-$16$) which contains a report of police seizing documents involving
specific names and dates.

%The user specifies documents $doc$-$029$ and $doc$-$026$ as the starting and ending documents. The initial story provided by the \textit{Storytelling Algorithm} by $A^*$ search is the sequence $doc$-$029\rightarrow doc$-$032 \rightarrow doc$-$026$. The user is not satisfied with the story and specifies that documents $doc$-$044$ and $doc$-$049$ should be on that story. Based on this feedback the \textit{Constrained $A^*$} search generates a new story $doc$-$029\rightarrow doc$-$044 \rightarrow doc$-$049 \rightarrow doc$-$016\rightarrow doc$-$026$. This is the shortest path $P^*$ between $doc$-$029$ and $doc$-$026$ in the new topic space. which includes the specified document $doc$-$044$ and $doc$-$049$ and brings a new document $doc$-$016$ which may be pertinent to connect the story in this new topic space.
%\vspace{-0.2cm}
\subsection{Evaluating story options}\label{sec:ks}
In this experiment, we seek to
generate multiple stories using our interactive storytelling approach as well as a vanilla
topic modeling, with a view to comparative evaluation. 
In this experiment, run over the Atlantic Storm dataset, the
user specifies $CIA06$ as the starting document and $NSA16$ as the ending document. 
The default story is:
$CIA06\rightarrow CIA37\rightarrow NSA19\rightarrow NSA16$. The user's feedback 
specifies $CIA08$ and $NSA09$ to be included in the final story. The results of
incorporating this feedback yields:
$CIA06\rightarrow CIA08\rightarrow DIA01\rightarrow NSA09\rightarrow NSA16$. We next
use Yen's $k$-shortest path algorithm~\cite{kShortest}
to generate a set of top $10$ (alternative) stories. As shown in
Table~\ref{table:story}, the top-ranked path in the interactive setting is indeed the
shortest path in the new topic space that satisfies the given constraints.
%\vspace{-0.1cm}
\begin{table*}[!t]
\centering
\scriptsize
\caption{Top $10$ stories (shortest paths) generated from $CIA06$ to $NSA16$ using both
a vanilla topic model and the interactive storytelling algorithm (using the Atlantic Storm
dataset).
The user's feedback requires that both $CIA08$ and $NSA09$ be included in the story. The interactive
storytelling algorithm updates the topic model wherein the shortest path indeed contains these documents.}
\label{table:story}
\begin{tabular}{r|c|r|c}
\hline %\hline
\multicolumn{1}{c|}{Top $10$ stories generated using vanilla topic model} & Path Length & \multicolumn{1}{c|}{Top $10$ stories generated using interactive storytelling} & Path Length \\ \hline
CIA06, CIA37, NSA19, NSA16                                        & 2.84        & CIA06, CIA08, DIA01, NSA09, NSA16                                     & 1.39        \\
CIA06, CIA20, CIA21, NSA16                                        & 3.16        & CIA06, CIA12, NSA09, NSA16                                            & 1.93        \\
CIA06, CIA22, CIA21, NSA16                                        & 3.16        & CIA06, CIA33, DIA01, NSA09, NSA16                                     & 2.13        \\
CIA06, CIA20, CIA22, CIA21, NSA16                                 & 3.16        & CIA06, CIA22, NSA09, NSA16                                            & 2.13        \\
CIA06, CIA22, CIA20, CIA21, NSA16                                 & 3.16        & CIA06, CIA08, DIA01, FBI07, NSA16                                     & 2.20        \\
CIA06, CIA08, NSA21, NSA16                                        & 3.23        & CIA06, CIA33, FBI07, NSA16                                            & 2.22        \\
CIA06, CIA08, NSA21, NSA12, NSA16                                 & 3.23        & CIA06, CIA33, CIA08, DIA01, NSA09, NSA16                              & 2.31        \\
CIA06, CIA08, NSA21, NSA13, NSA16                                 & 3.23        & CIA06, CIA11, FBI13, DIA01, NSA09, NSA16                              & 2.33        \\
CIA06, CIA08, NSA21, NSA12, NSA13, NSA16                          & 3.23        & CIA06, DIA02, DIA01, NSA09, NSA16                                     & 2.33        \\
CIA06, CIA08, NSA21, NSA18, NSA16                                 & 3.23        & CIA06, CIA08, CIA23, NSA16                                            & 2.34        \\ \hline %\hline
\end{tabular}
\end{table*}
 %\vspace{-0.1cm}
%\vspace{-0.05cm}
\subsection{Proximity between topics}\label{sec:proT}
\begin{figure*}[!t]
  \centering
  \begin{tabular}{@{}c@{}c@{}c@{}}
  \includegraphics[width=0.3\textwidth,height=1.8in]{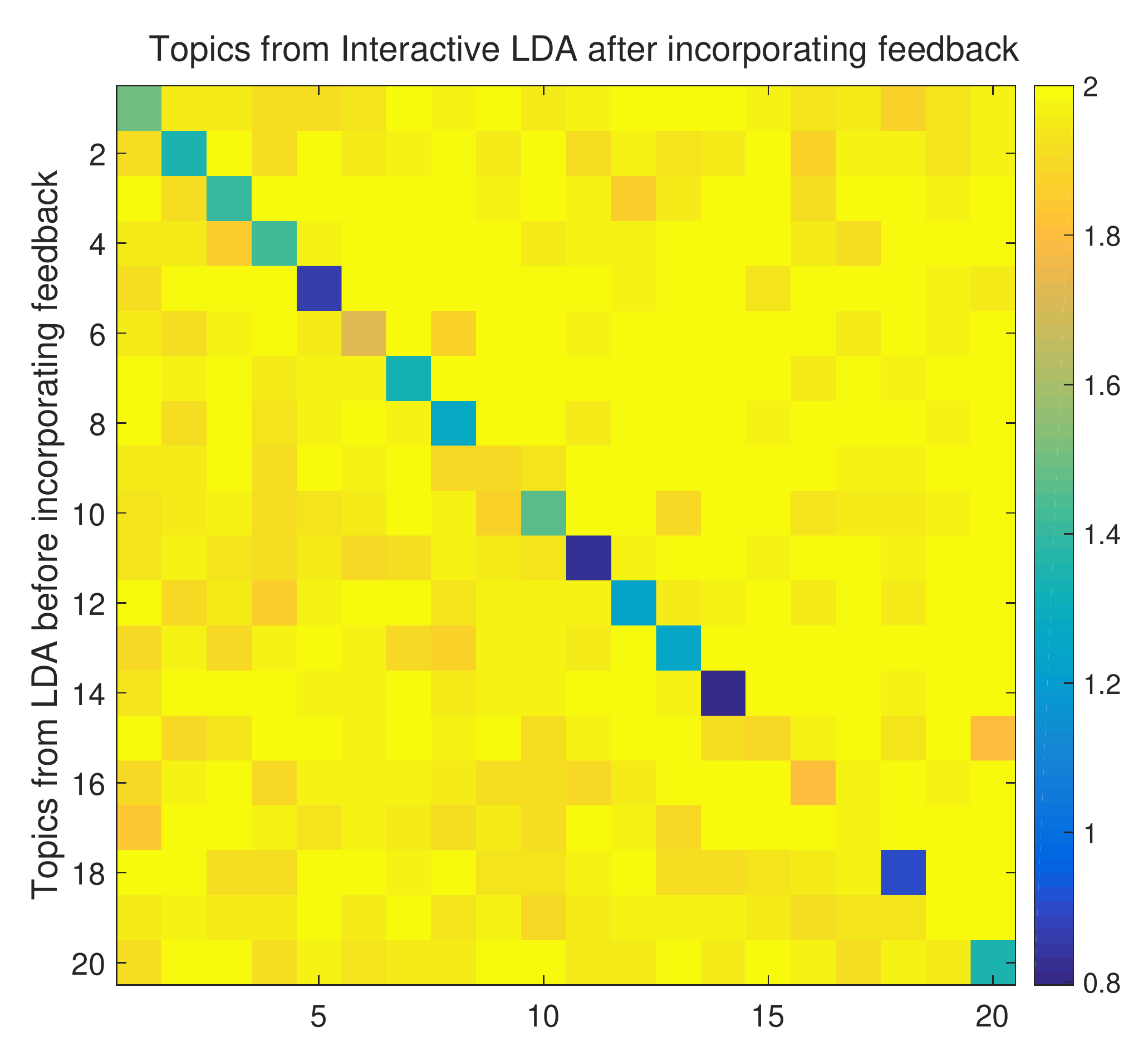} & 
  \includegraphics[width=0.3\textwidth,height=1.8in]{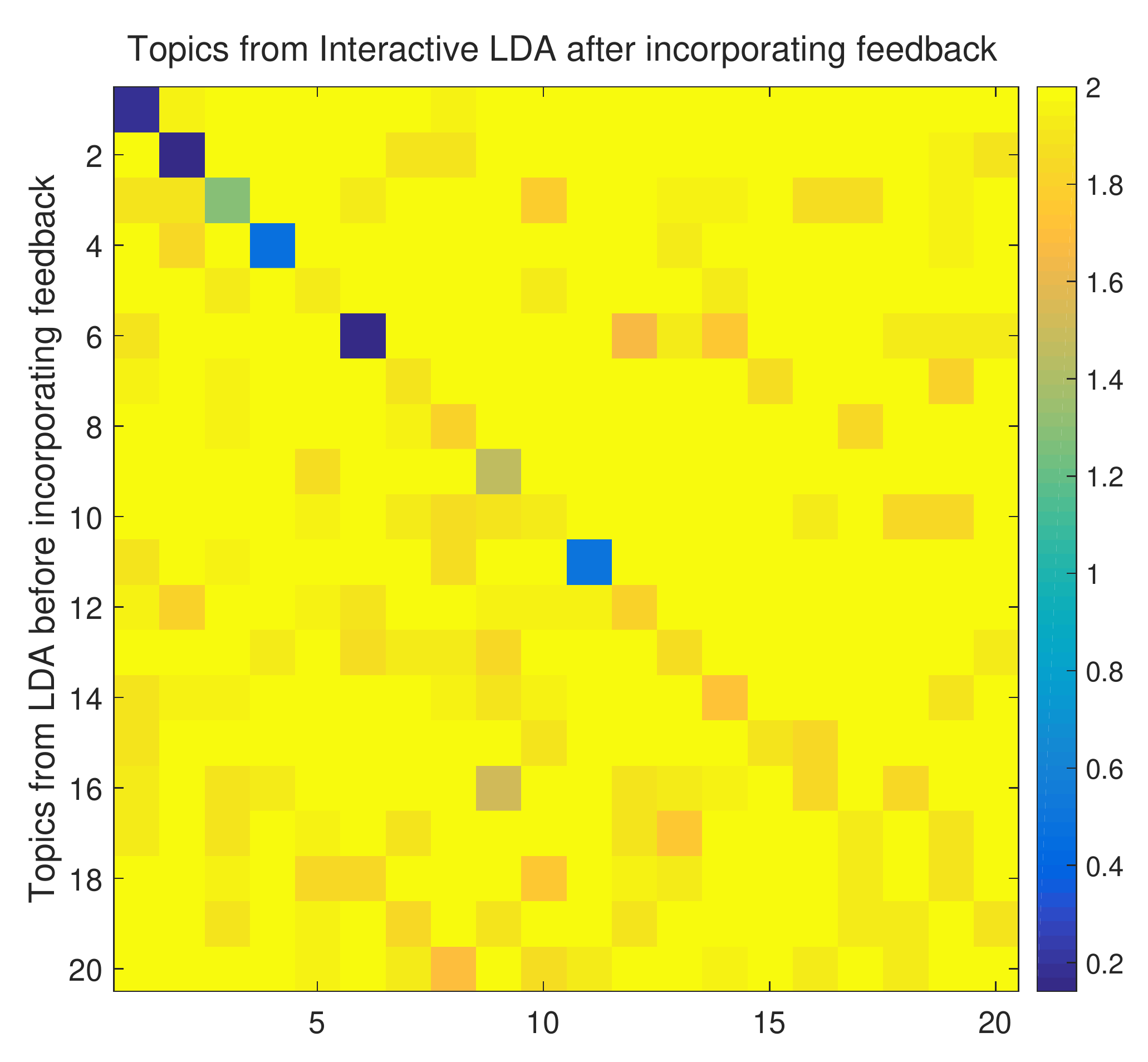} &
  \includegraphics[width=0.3\textwidth,height=1.8in]{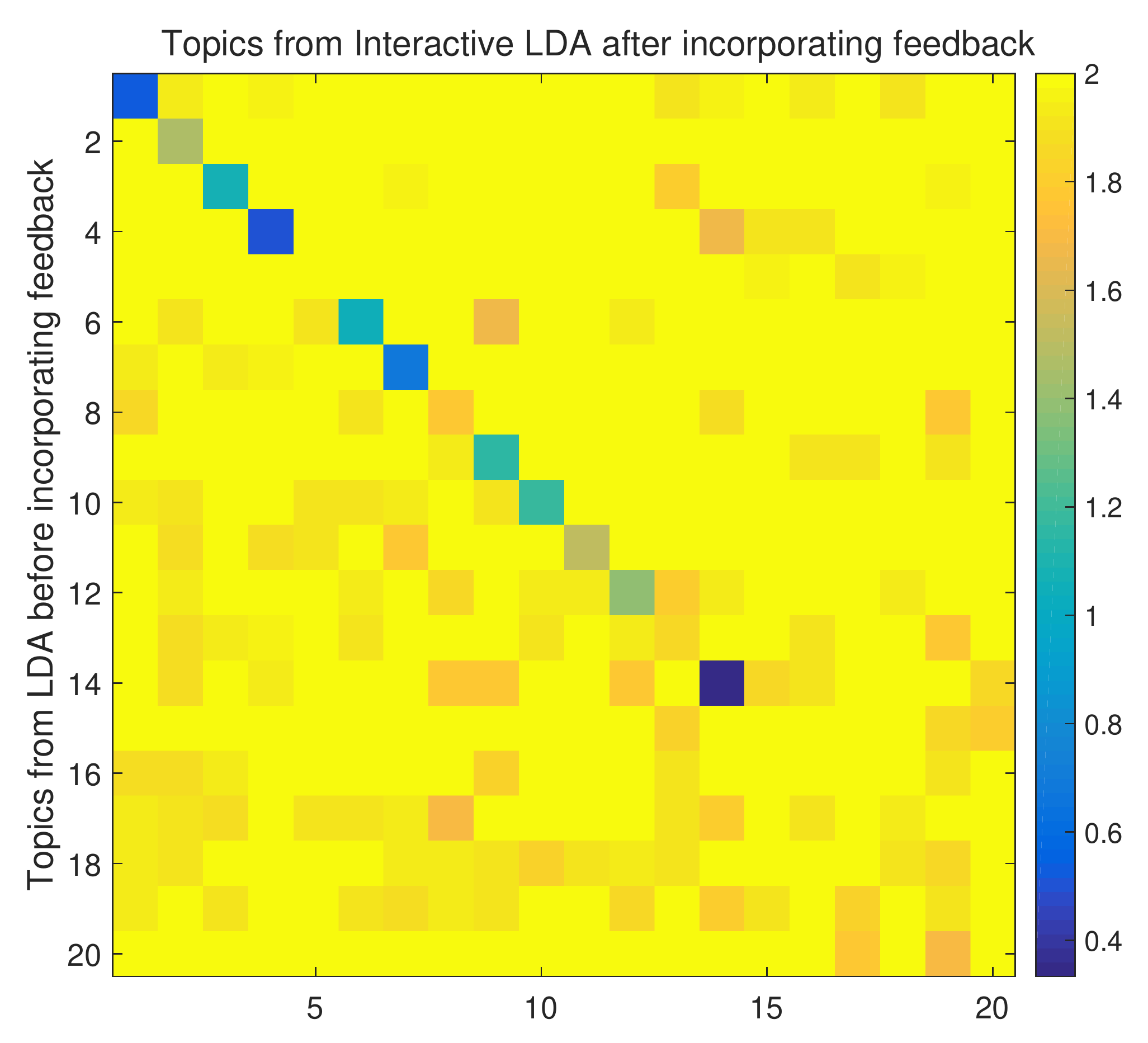} \\
  Atlantic Storm & Crescent & Manpad \\  
  \end{tabular}
  \caption{Manhattan distance between topic distributions before and after user feedback. Blue color denotes topics closest to each other. As can be seen, the incorporation of feedback destroys the diagonal dominance of the matrix.}
  \label{fig:heat}
\end{figure*}

We investigate topic proximity in terms of Manhattan distance in Fig~ \ref{fig:heat}. Here,
rows denote topics from a vanilla 
topic model, and the columns correspond to topics inferred by the interactive storytelling algorithm.
As shown in Fig.~\ref{fig:heat} the diagonally dominant nature of the matrix is destroyed due to the introduction
of user feedback, illustrating that the distributions of words underyling the topics are quite dissimilar.

\iffalse
 shows the distance between a pair of topics from original LDA while the column shows the distance from LDA \textit{Interactive Storytelling}. The closest \textit{Interactive Storytelling} based topic for each LDA is shown in the diagonal. We can see that topics redefinition after incoporting feedback are quite different from the initial topics from vanilla LDA. Since topics are represented in terms of distribution the highest distance between two topics is 2 units. We can see that some of the distances along the diagonal are closer to 2. We also show top four terms from each computed by the both topic models in Table \ref{table:topLDA} and \ref{table:topILDA} for Atlantic Storm dataset. These terms also show that the topic redefinitions of two topic models are quite different from each other.
\fi
%\begin{figure}[hbtp]
%\centering
%\includegraphics[width=0.5\textwidth]{fig/heat_map.pdf}
%\caption{Manhattan distance between the topic distribution from two topic models in Atlantic Storm dataset. The row corresponds to the Manhattan distance between a pair of topics from vanilla LDA (prior feedback) and column corresponds to the same metric between a pair of topics from \textit{Interactive LDA} (post feedback). Each LDA based topic has been aligned with closet \textit{Interactive LDA} based topics along the diagonal. Blue color means topics are closer to each other.}
%\label{fig:heat}
%\end{figure}
\subsection{Scalability to large corpora}\label{sec:clusS}
\begin{figure*}[!t]
\centering
\includegraphics[width=0.95\textwidth]{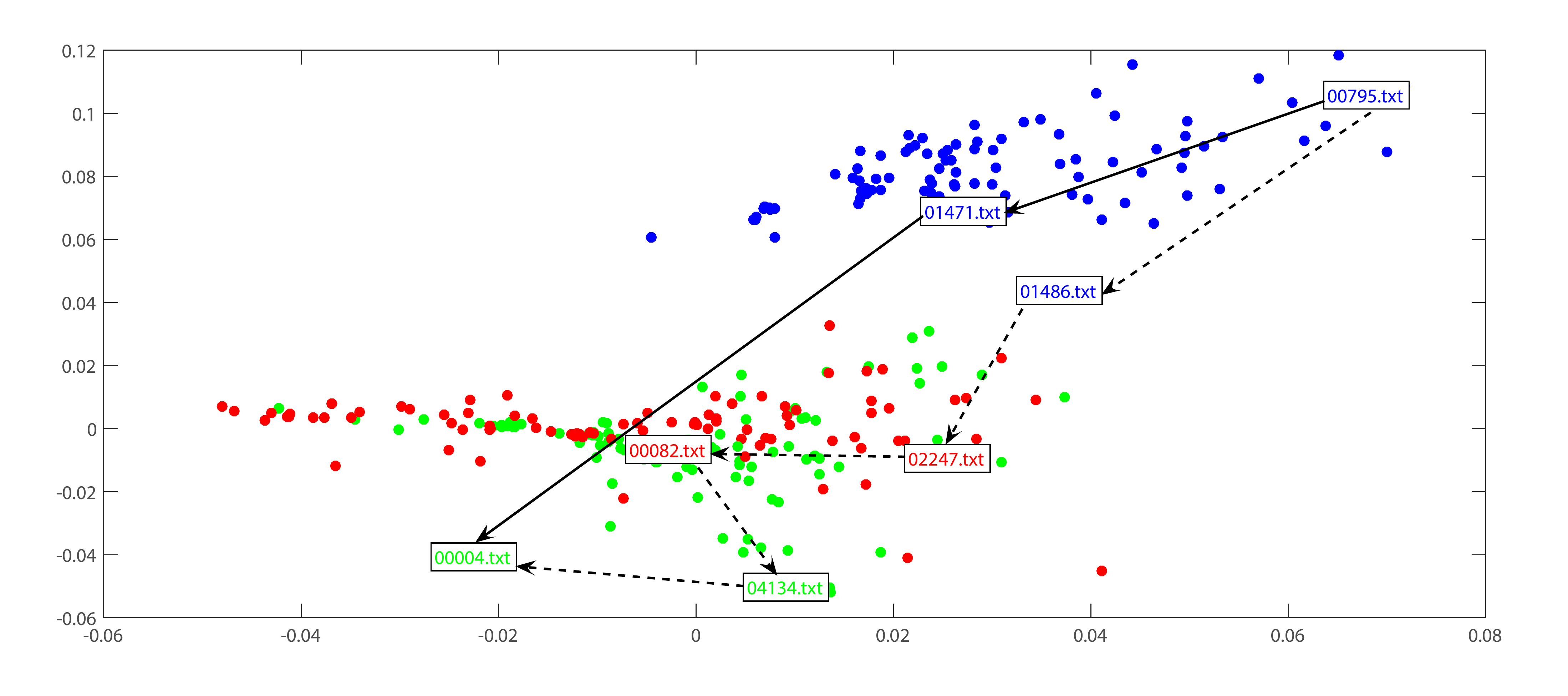}
\caption{Scaling the storytelling methodology by integrating clustering. The initial story (solid line) from
$00795.txt$ to $00004.txt$ avoided documents in the red cluster. After incorporating user feedback, the new
story (dotted line) navigates through the red cluster.}
\label{fig:mds_vast_clu}
\end{figure*}

With large datasets, such as the VAST11 dataset, we can fruitfully combine clustering with our
framework to navigate the document collection (see Fig.~\ref{fig:mds_vast_clu}).
Given a document collection, an initial clustering (e.g., k-means) can be utilized to identify broad groups of documents that
can be discarded during the initial story construction.
Here, assume that the user specifies $00795.txt$ and $00004.txt$ as the starting and ending document, respectively. 
The storytelling algorithm
generates $00795.txt \rightarrow 014171.txt \rightarrow 00004.txt$ as the initial story (solid line). 
Note that this story ignores documents from the cluster displayed in red. Assume that 
the user now requires that documents from 
the red cluster also participate in the story. Based on an initial exploratory analysis, the user
specifies that  
documents $02247.txt$ and $00082.txt$ should participate in
the story. Based on this feedback the interactive
storytelling algorithm generates:
$00795.txt\rightarrow 01486.txt\rightarrow 02247.txt\rightarrow 00082.txt\rightarrow 04134.txt\rightarrow 00004.txt$
(note the introduction of $04134.txt$ into the story).

%\vspace{-0.1cm}
\subsection{Comparing interactive storytelling vs uniform cost search}\label{sec:com}
\begin{figure*}[!t]
  \centering
  \begin{tabular}{@{}c@{}c@{}c@{}}
  \includegraphics[width=0.3\textwidth,height=1.7in]{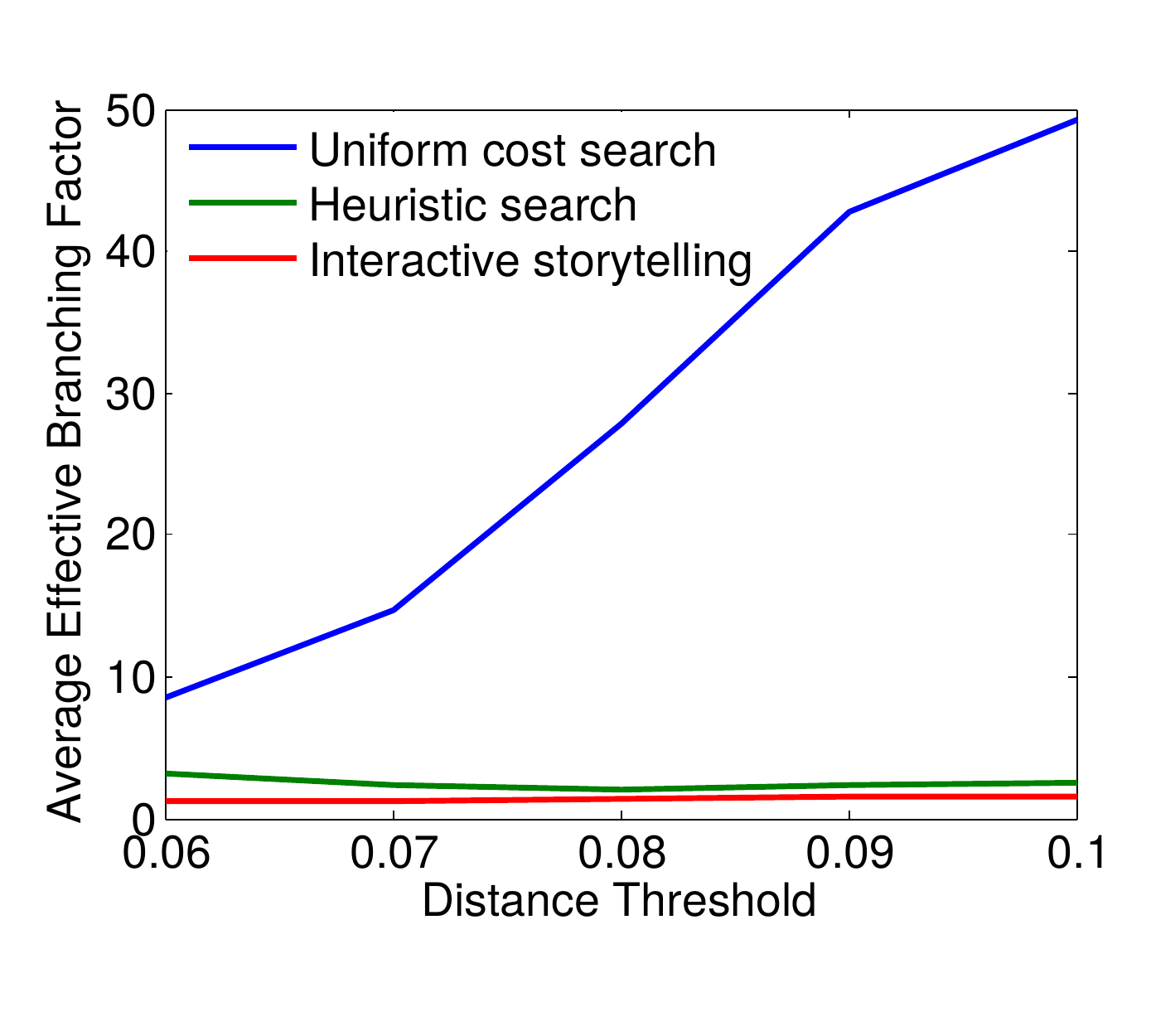} & 
  \includegraphics[width=0.3\textwidth,height=1.7in]{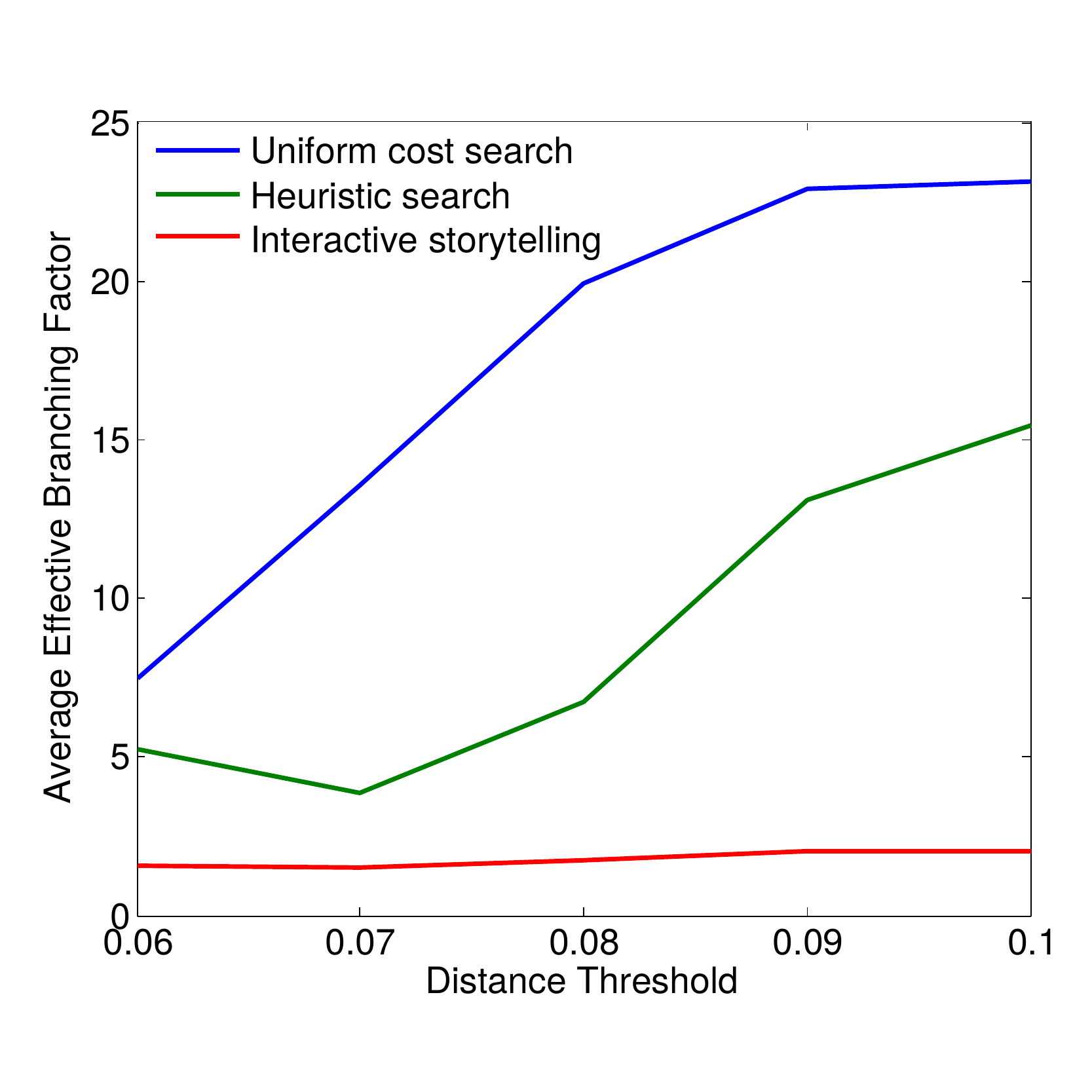} &
  \includegraphics[width=0.3\textwidth,height=1.7in]{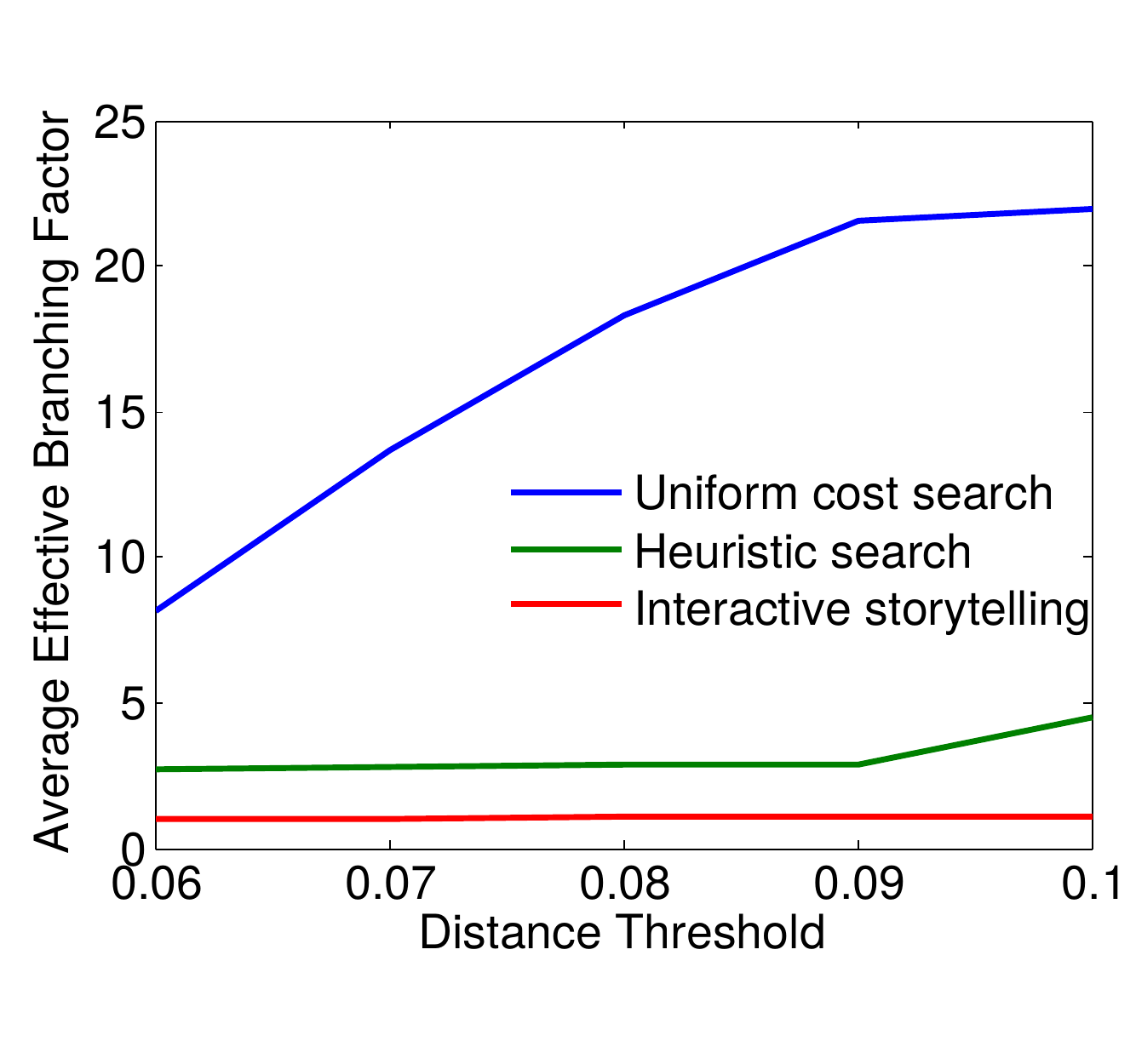} \\
  (a) & (b) & (c) \\  
  \includegraphics[width=0.3\textwidth,height=1.7in]{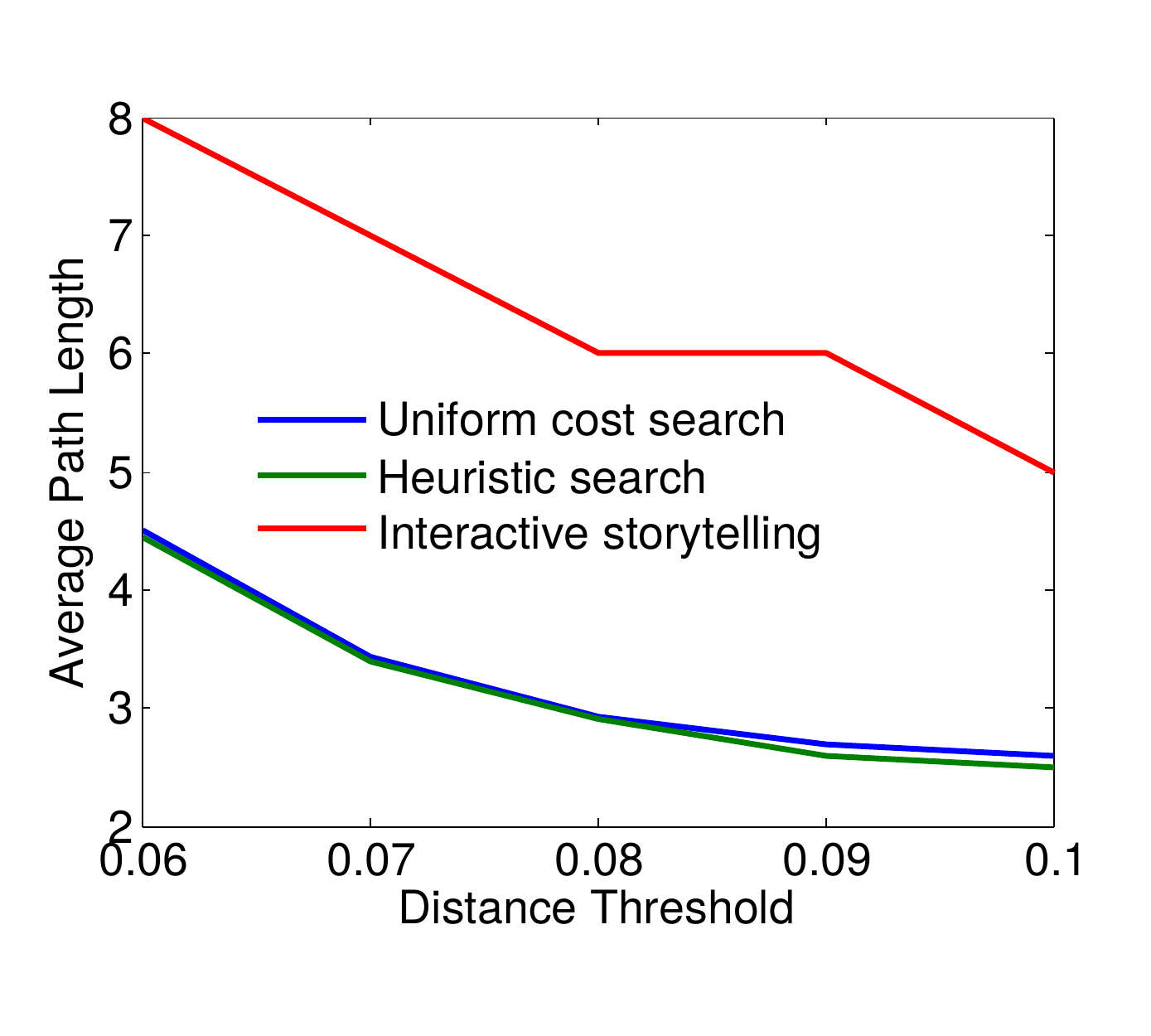} & 
  \includegraphics[width=0.3\textwidth,height=1.7in]{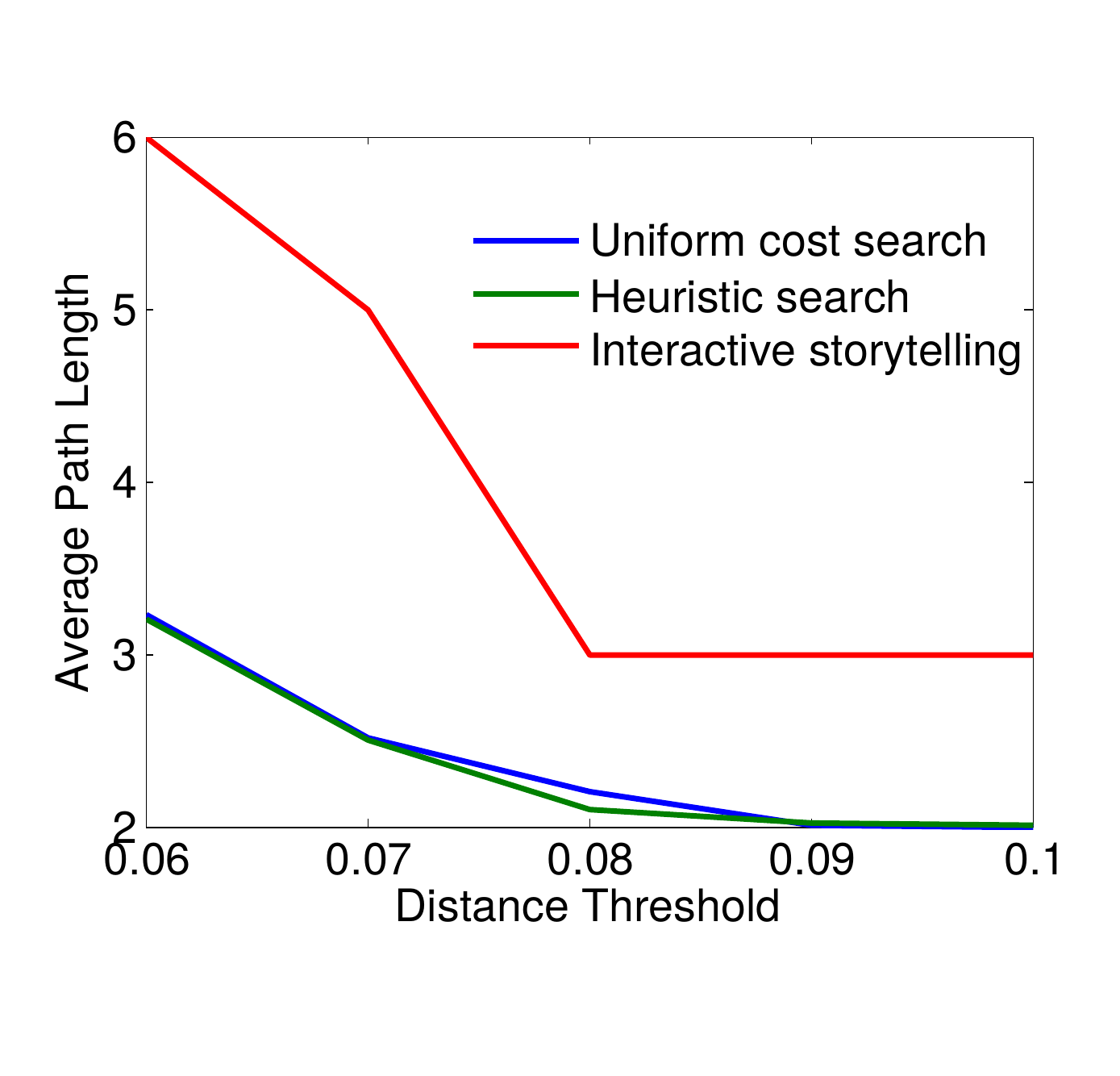} &
  \includegraphics[width=0.3\textwidth,height=1.7in]{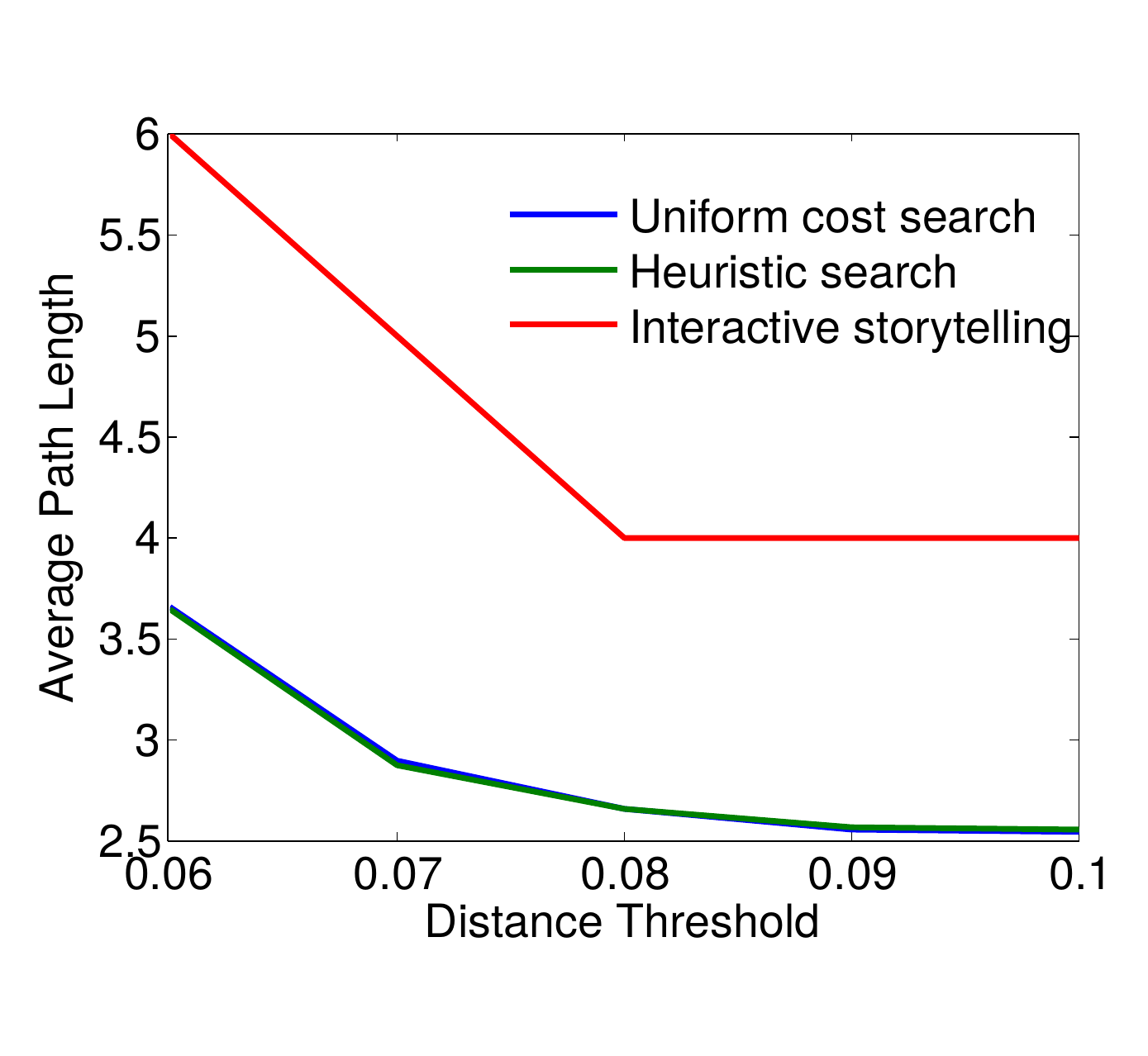} \\
  (d) & (e) & (f) \\  
  \includegraphics[width=0.3\textwidth,height=1.7in]{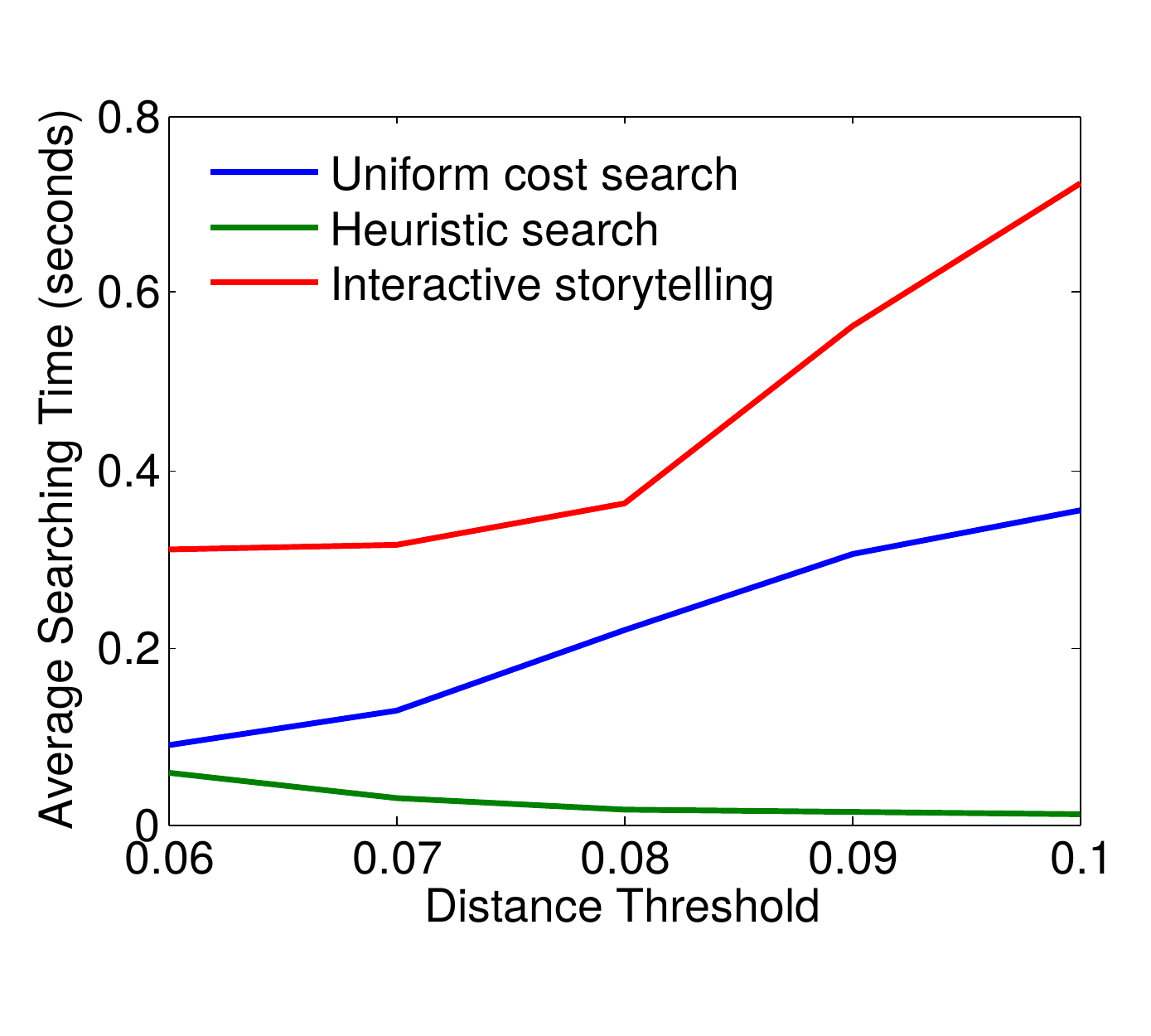} & 
  \includegraphics[width=0.3\textwidth,height=1.7in]{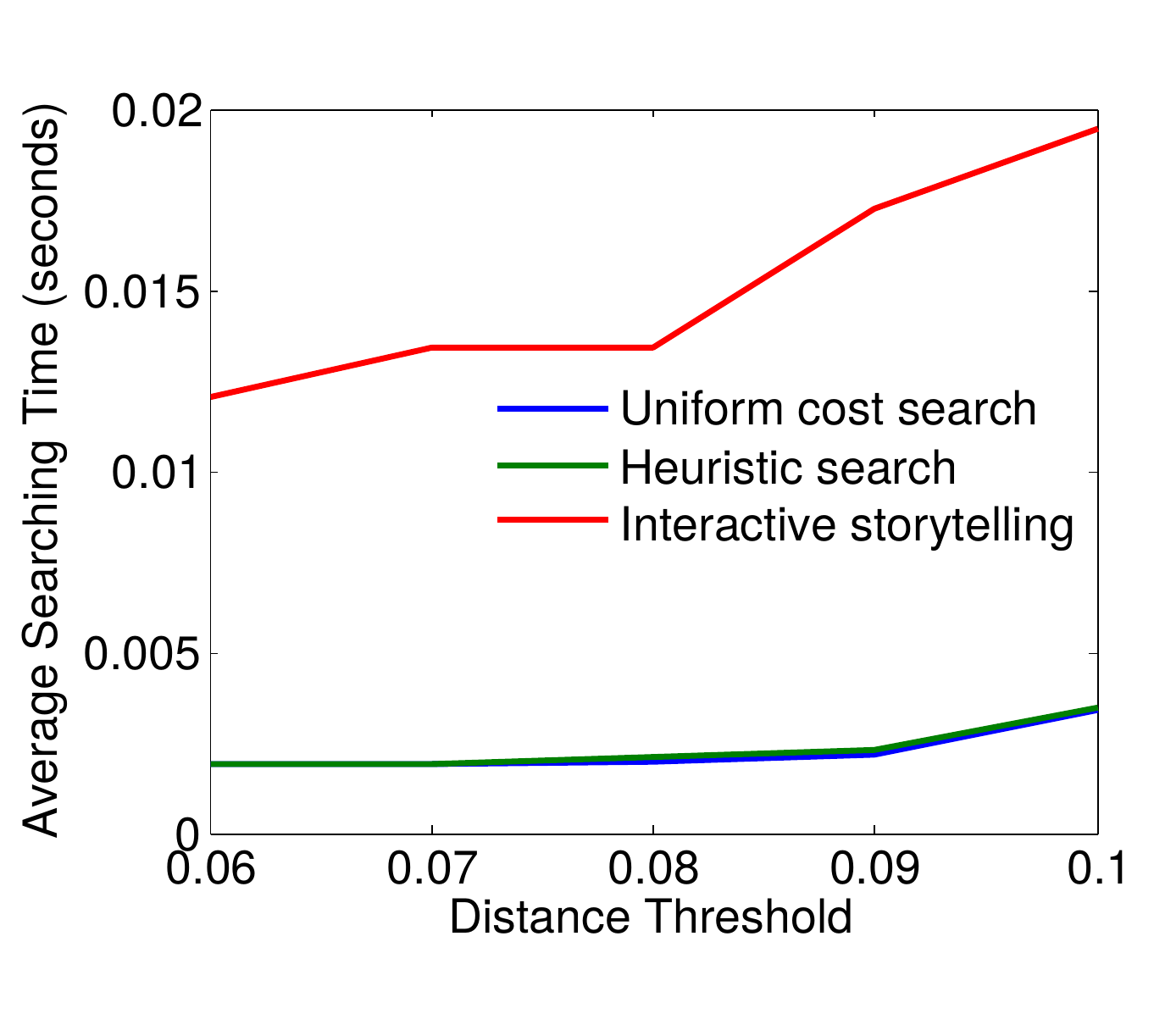} &
  \includegraphics[width=0.3\textwidth,height=1.7in]{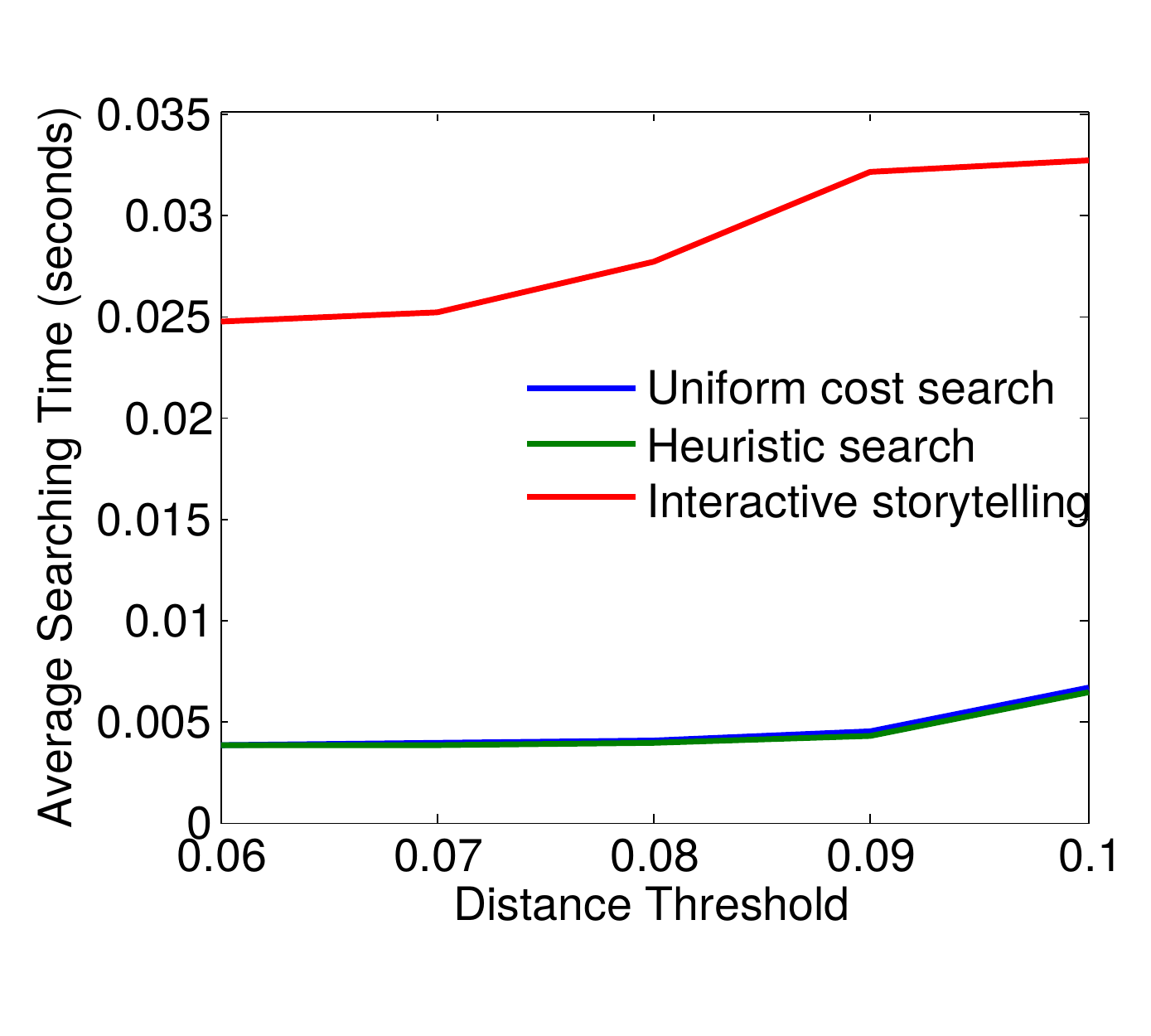} \\
  (g) & (h) & (i) \\
  \end{tabular}
  \caption{Comparison of interactive storytelling, heuristic search and uniform cost search in terms of average effective branching factor (top), average path length (middle) and execution time (bottom). (Left) Atlantic Storm. (middle) Crescent. (right) Manpad.}
  \label{fig:comp}
\end{figure*}

We now assess the performance of the constrained search process underlying
interactive storytelling versus that of an uninformed search (e.g., uniform cost search). The comparison is shown in Fig. \ref{fig:comp}. We use different distance threshold $\xi$ to compute effective branching factor, path length and execution time.%Figs %\textit{Interactive Storytelling} in terms of average effective branching factor, average path length and computation time. 

We show in Fig. \ref{fig:comp}(a, b, c) that average effective branching factor increases with $\xi$. Since higher $\xi$ means a node will have more neighbors, the branching factor will increase in this case. However in case of \textit{Interactive Storytelling} path finding is more guided so the average effective branching factor does not vary much. We can see that using a heuristics decreases the average effective branching factor. The average path length however decreases with increasing $\xi$ (Fig. \ref{fig:comp}(d, e, f)). Increasing $\xi$ results in having larger neighborhood for each node, therefore the chance of reaching the goal becomes higher resulting in shorter average path length. For \textit{Interactive Storytelling} the average path length is higher because it has to visit the nodes specified by the user while searching for the shortest path. The execution time for both heuristic search and the uninformed search are almost same (Fig. \ref{fig:comp}(g, h, i)), however for \textit{Interactive Storytelling} it is much longer. Since it has to visit the nodes provided by the user, it travels the search space in more depth so it takes more time on average to finish the search.

\renewcommand{\arraystretch}{0.1}
%\vspace{-0.2cm}
\section{Related Work}
%In this section we discuss related work to topic modeling. 
Related work pertaining to storytelling has been covered in the introduction. We survey topic modeling
related work here. To the best of our knowledge, no existing work supports the incorporation of path-based
constraints to refine topic models, as done here.
%\vspace{-0.3cm} 
%Topic modeling algorithms are now widely established in data mining practice. Building upon the earlier works like  pLSA~\cite{plsa} methods such as LDA~\cite{lda} represent documents as a mixture of topics where each topic is a distribution over words in a vocabulary. Many variants of LDA have been proposed in the literature. We have observed recent development in topic modeling has been progressed along three lines.
%
\paragraph{Expressive topic models}
The author-topic model~\cite{atm} is one of the popular extensions of topic models
that aims to model how multiple authors contributed to a document collection.
Works such as~\cite{genSpcePTM,combSem} extend basic topic
modeling to include specific words or 
semantic concepts by incorporating
notions of proximity between documents. In~\cite{tmbbow}, the authors move beyond 
bag-of-words assumptions and accommodate the  ordering of words in topic modeling. Domain
knowledge is incorporated in~\cite{domTM} in the form of
Dirichlet forest priors. Finally, in~\cite{ctm}, correlated topic models are introduced to model correlations
between topics.
%\vspace{-0.3cm} 
\paragraph{Incorporating external information}
Supervised topic models are introduced in~\cite{slda}.
Lu and Zhai~\cite{opInTM} propose a semi-supervised topic model to incorporate 
expert opinions into modeling. In~\cite{llda}, authors incorporate user tags accorded to documents to
place constraints on topic inference.
The timestamps of documents is used in~\cite{dtm,tot}
to model the evolution of topics in large corpus. 
%\vspace{-0.3cm} 
\paragraph{Visualizing topics}
Wei et al.~\cite{tiara} propose TIARA, a visual exploratory text analytics system to observe the 
evolution of topics over time in a corpus. Crossno et al.~\cite{tv} develop a framework to visually compare 
document contents based on different topic modeling approaches. In~\cite{tpNet}, the
authors present documents in topic space and depict inter-document connectivity as a network in a visual interface,
simultaneously displaying community clustering.
%\vspace{-0.3cm}
\paragraph{Interactive topic modeling}
User feedback is incorporated in~\cite{itm} wherein users can provide constraints about specific words that
must appear in topics. An active learning framework to incorporate user feedback and improve topic quality is introduced
in~\cite{alCTM}.

%\vspace{-0.2cm}
\section{Discussion}
We have demonstrated interactive storytelling, a combination of interactive topic modeling and
constrained search wherein documents are connected obeying user constraints on paths. User feedback is
pushed deep into the computational pipeline and used to refine the topic model. Through experiments we
have demonstrated the ability of our approach to provide meaningful alternative stories while satisfying
user constraints. In future work, we aim to generalize our framework to a multimodal network representation
where entities of various kinds are linked through a document corpus, so that constraints can be more
expressively communicated.

%\bibliographystyle{abbrv}
%\bibliography{refe}

\end{document}